\documentclass{agujournal2019}
\usepackage{lmodern}
\usepackage{xcolor}
\usepackage{lscape}
\usepackage{longtable}
\usepackage{array}
\usepackage{amsmath}
\usepackage{amssymb}
\usepackage{multirow}
\usepackage{soul}
\usepackage{lineno}
\usepackage{mathtools}
\usepackage{float}
\usepackage{comment}

\begin{document}

\graphicspath{
               {edited_figures/}
              }

\title{Deep Learning for Simultaneous Inference of Hydraulic and Transport Properties}
\authors{Zitong Zhou\affil{1}, Nicholas Zabaras\affil{2}, and Daniel M. Tartakovsky\affil{1}}

\affiliation{1}{Department of Energy Resources Engineering, Stanford University, Stanford, CA 94305, USA}

\affiliation{2}{Scientific Computing and Artificial Intelligence (SCAI) Laboratory, University of Notre Dame, South Bend, IN 46556, USA}

\correspondingauthor{Daniel M. Tartakovsky}{tartakovsky@stanford.edu}
\correspondingauthor{Nicholas Zabaras}{nzabaras@gmail.com}

%%%%%%%%%%%%%%%%%%%%% KEYPOINTS %%%%%%%%%%%%%%%%%%%%%%%%%%%
\begin{keypoints}
\item We present a deep-learning strategy to reconstruct conductivity and contaminant release history in three-dimensional heterogeneous aquifers.
\item Conductivity parameterization with convolutional adversarial autoencoder reduces the inverse problem's dimensionality.
\item Convolutional encoder-decoder acts as a surrogate of forward models; ensemble smoother approximates parameters' posterior distribution.
%\item Comparison to the physical model based inversion framework CAAE-ESMDA shows that our CAAE-DenseED-ESMDA achieves an accurate and efficient solution.
\end{keypoints}

%%%%%%%%%%%%%%%%%%%%%% ABSTRACT %%%%%%%%%%%%%%%%%%%%%%%%%%%
\begin{abstract}
Identification of a heterogeneous conductivity field and reconstruction of a contaminant release history are  key aspects of subsurface remediation. These two goals are achieved by combining model predictions with sparse and noisy hydraulic head and concentration measurements. Solution of this inverse problem is notoriously difficult due to, in part, high dimensionality of the parameter space and high computational cost of repeated forward solves. We use a convolutional adversarial autoencoder (CAAE) to  parameterize a heterogeneous non-Gaussian conductivity field via a low-dimensional latent representation. A three-dimensional dense convolutional encoder-decoder (DenseED) network serves as a forward surrogate of the flow and transport model.  The CAAE-DenseED surrogate is \textcolor{blue}{fed} into the ensemble smoother with multiple data assimilation (ESMDA) algorithm to sample from the Bayesian posterior distribution of the unknown parameters, forming a CAAE-DenseED-ESMDA inversion framework. The resulting CAAE-DenseED-ESMDA inversion strategy is used to identify a three-dimensional contaminant source and conductivity field. A comparison of the inversion results from CAAE-ESMDA with physical flow and transport simulator and from CAAE-DenseED-ESMDA shows that the latter yields accurate reconstruction results at the fraction of the computational cost of the former. 
\end{abstract}
% \linenumbers

%%%%%%%%%%%%%%%%%%%%%% I. INTRODUCTION %%%%%%%%%%%%%%%%%%%%%%%%%%%
\section{Introduction}\label{intro}

Design of regulatory and remedial actions for contaminated soils and aquifers rely on reconstruction of the contaminant release history. Given subsurface heterogeneity, this task is inseparable from the need to identify hydraulic and transport properties of the subsurface environment. Both tasks have to contend with sparse and noisy measurements collected many years or decades after the contamination event took place. Prior to recent breakthroughs in computer architecture and algorithmic development, this joint inversion of hydraulic and water-quality data for real-world problems was so computationally demanding as to defy a solution unless dramatic (and often unrealistic) simplifications of the problem were made. For example, past efforts to reconstruct a contaminant release history found it necessary to assume solute migration to be one-or two-dimensional and subsurface properties, such as hydraulic conductivity $K(\mathbf x)$, to be known with certainty~\cite[among many others]{aral2001identification, snodgrass1997geostatistical, yeh2007groundwater}. Yet, aquifers are seldom, if ever, homogeneous, with $K(\mathbf x)$ often varying by orders of magnitude within the same aquifer and exhibiting highly non-Gaussian, multimodal behavior~\cite{tartakovsky-2008-uncertain, winter-2003-moment, yang-2020-method}. Likewise, while the assumption of two-dimensional groundwater flow is often valid, accounting for the three-dimensional nature of contaminant migration is essential to prediction accuracy.

Our effort in joint inversion of hydraulic conductivity and contaminant release history from error-prone measurements of  hydraulic head and solute concentration revolves around two challenges. The first is to  describe the unknown non-Gaussian heterogeneous conductivity field with an adequate prior distribution. The second is to estimate a large number of unknown parameters in the inverse problem.

To tackle the first challenge, a parameterization of the high-dimensional conductivity field with a low-dimensional latent variable is commonly used~\cite{linde2015geological, zhou2014inverse}. Parameterizations based on the principle component analysis (PCA)~\cite{sarma2008kernel, vo2014new} perform well for Gaussian random fields, but require ad-hoc modifications for non-Gaussian fields~\cite{liu2019deep}. DNN-based parameterizations  eliminate the need for the Gaussianity assumption~\cite{canchumuni2019history, liu2019deep}. Two popular  methods of this class are the generative adversarial network (GAN)~\cite{goodfellow2014generative} and variational autoencoders~\cite{kingma2013auto}. Both produce a DNN that learns a two-way mapping between the conductivity field and a low-dimensional embedding. Random realizations from the latent variable distribution can be decoded to a conductivity field that is statistically similar to those drawn from the training data set. A series of studies involving usage of variational autoencoders or GANs in~\cite{ laloy2017inversion, laloy2018training, laloy2019gradient, lopez2021deep} have shown their superiority over PCA for inversion tasks in geophysics, specifically in geophysical formation exhibiting channel patterns. In addition to the large reduction in the number of the unknown parameters, such parameterizations make it  feasible to tackle the latent variable distribution, which is typically a standard normal by construction. This simplicity, in turn, facilitates the solution of the inverse problem with ensemble methods discussed below. 

The second challenge, high-dimensionality of the parameter space, manifests itself in significant computational burden of an inversion procedure.  Parameter estimation, which lies at the heart of an inverse problem, is achieved by matching the noisy measurements with the prediction of a flow and solute transport model. Strategies for solving typical ill-posed inverse problems fall into two main categories, deterministic and probabilistic. Deterministic methods, such as least square regression~\cite{white2015nonlinear} and hybrid optimization with a genetic algorithm~\cite{ayvaz2016hybrid, leichombam2018new}, seek a ``best'' estimate of the unknown parameters, without quantifying the  uncertainty inherent in this type of problems. Probabilistic methods, such as Markov Chain Monte Carlo or MCMC~\cite{gamerman2006markov} and data assimilation via Kalman filters~\cite{evensen1994sequential, evensen2003ensemble, xu2016joint, xu2018simultaneous} and their variants~\cite{Emerick2013,zhang2018iterative}, overcome this shortcoming of their deterministic counterparts. Yet, the high-cost of necessary repeated forward solves undermines their utility  for large, complex inverse problems, unless dedicated high-performance computing facilities are available for the task.  

Two complementary strategies can be deployed to alleviate this cost. The first aims to reduce the number of forward simulations needed for an inversion algorithm to converge. The second seeks to reduce the computational cost of each forward solve. \textcolor{blue}{We divide the discussion about the former direction into two parts: when using MCMC as the inference method; and when ensemble data assimilation methods are explored instead of MCMC. 1. Design and application of efficient MCMC variants has been an active research area in the past two decades. The most popular variant of non-gradient-based MCMC might be }delayed rejection adaptive Metropolis (DRAM) sampling~\cite{haario2001adaptive, haario2006dram} which slightly outperforms a random walk Metropolis-Hastings MCMC in terms of efficiency~\cite{zhang2015efficient, zhou2021markov, xia2021bayesian}. Gradient-based MCMC methods, such as hybrid Monte Carlo (HMC) sampling~\cite{barajas2019efficient}, converge  faster than these and other MCMC variants. However, computation of the gradient of a Hamiltonian dynamical system is prohibitive for high-dimensional transport problems. Learning on statistical manifolds provides another possible solution~\cite{boso2020learning, boso2020data}. \textcolor{blue}{Given the various exploration directions on MCMC, MCMC methods are still considered expensive, as they are not feasible for parallelization~\cite{ghorbanidehno2020recent}given the requirement for the Markov chain to reach an equilibrium state. On the other hand,} Ensemble-based inversion methods are generally faster since they allow nearly perfect parallelization, because of the independence of samples in the ensemble. Variants of Kalman filters, such as iterative Ensemble Kalman filter (IEnKF), have been used for estimation of three-dimensional heterogeneous permeability fields~\cite{chaudhuri2018iterative}. \textcolor{blue}{In principle, any Kalman filter based ensemble methods can be applied to solve inverse problems like in this study. Particularly,} a relatively new variant of Kalman filter, ensemble smoother with multiple data assimilation (ESMDA)~\cite{Emerick2013}, has gained popularity in subsurface flow history matching~\cite{tang2021deep,kim2019history,jiang2021data}. Originally developed as a decision making tool in the fields of energy efficiency and petroleum engineering, Bayesian evidential learning has evolved rapidly and shown its potential in other applications~\cite{hermans2018uncertainty,hermans2019bayesian, michel20201d}. Subsequent efforts adapted this method to the tasks of parameter estimation and optimal design~\cite{thibaut2021new, park2020direct}. We posit that ESMDA is an efficient tool for solving high-dimensional inverse problems with sparse and noisy observations.

In terms of inversion complexity, we subdivide recent groundwater-related studies into three categories: estimation of hydraulic conductivity from measurements of hydraulic head and, optionally, of solute concentration~\cite{mo2019integration, ju2018adaptive}; estimation of contaminant release history from concentration measurements, for known flow and transport parameters~\cite{zhou2021markov, zhang2015efficient}; and estimation of both contaminant release history and hydraulic conductivity from hydraulic head and solute concentration data in two-~\cite{mo2019deep, xu2018simultaneous, kang2021hydrogeophysical} and three-dimensional~\cite{kang2020improved} aquifers. We briefly discuss the latter category to highlight the novelty of our approach. 

A low-dimensional representation of the random log-normal conductivity obtained via the Karhunen-Lo\`eve expansion (KLE) \cite{mo2019deep} loses its attractiveness if the subsurface environment is highly  heterogeneous, exhibiting short correlation lengths and multimodal statistics. Our work is among the few recent   source identification studies dealing with nonlinear transport problems and nonstationary (statistically inhomogeneous) conductivity fields. The deep learning-based strategies of ensemble inversion were adopted in~\cite{xu2018simultaneous, kang2021hydrogeophysical} to estimate both a non-Gaussian conductivity field and the source of contamination, yet they deal only with two-dimensional problems, and their accuracy is relatively low. In the adjacent field of petroleum engineering, CNN post-processing of PCA (CNN-PCA) parameterization and ESMDA were used to estimate both a channelized permeability and oil/water rate~\cite{tang2021deep}. However, this application deals with an observable quantity (oil/water production rate), while ours has to contend with an unobservable one  (the location and strength of a contaminant release). \textcolor{blue}{In more words, data assimilation in petroleum engineering aims to match the observed oil/water production rate with the inferred permeability field and unknown injection history, rather than obtaining accurate permeability field itself like in hydrology studies. In groundwater studies, the unknown contaminant source terms are unobservable, hence the "calibration" goal in petroleum engineering will not be sufficient in hydrology field.}

\textcolor{blue}{The} shortcomings of the joint inversion strategies mentioned above \textcolor{blue}{can be summarized as these several aspects: limitation of parameterization of non-Gaussian or channelized field; unclear demonstration of applicability on high dimensional inverse problems; uncertain generalization to inverse problems emphasized on the unobservable history. To address these shortcomings}, we use a convolutional adversarial autoencoder (CAAE) to parameterize a non-Gaussian conductivity field~\cite{mo2019integration}, train a surrogate dense encoder-decoder DNN to replace the PDE-based model of subsurface flow and transport, and apply the ESMDA inversion framework to identify the spatiotemporally extended source of contamination and the latent variables representing the conductivity field.  We posit that combination of these three components, which yields the method we refer to as CAAE--DenseED--ESMDA, provides a fast and robust inversion solution.  Although advantages of each component were established in various disciplines, their synergy has remained unexplored. Our study demonstrates that CAAE--DenseED--ESMDA is a powerful tool for dynamic source identification and conductivity reconstruction in three spatial dimensions, when the number of unknown parameters is large ($\sim 20000$). \textcolor{blue}{\st{We are not aware of other machine learning strategies used to tackle inverse groundwater problems of similar complexity.}}

In Section~\ref{sec:prbform}, we formulate the problem of joint reconstruction of hydraulic conductivity field and contaminant release history from sparse and noisy measurements of hydraulic head and solute concentration. Our inversion strategy, combining CAAE parameterization of the conductivity field (Section~\ref{sec:CAAE}), a convolutional DNN surrogate of the flow and transport model (Section~\ref{sec:cnn}), and the ESMDA inversion method (Section~\ref{sec:ESMDA}), is described in Section~\ref{sec:methods}. Results of our numerical experiments are reported in Section~\ref{sec:results}; they demonstrate that our method is about $8$ times faster than CAAE-ESMDA with the PDE-based flow and transport model. Main conclusions drawn from this study are summarized in Section~\ref{sec:concl}. Details of the neural network architectures are given in Supporting Information (SI).

%%%%%%%%%%%%%%%%%%%%%% II. Problem formulation %%%%%%%%%%%%%%%%%%%%%%%%%%%
\section{Problem Formulation}
\label{sec:prbform}

The problem formulation consists of the description of a reactive transport model (Section~\ref{sec:lastPDEs}) and the specification of a data model (Section~\ref{sec:lastData}).

\subsection{Contaminant Transport Model}
\label{sec:lastPDEs}

We consider transport of a reactive solute in a three-dimensional steady-state groundwater flow field. The latter is described by:
\begin{linenomath*}
\begin{equation}
    \nabla \cdot (K \nabla h) = 0, \qquad \mathbf x = (x_1,x_2,x_3)^\top \in \Omega \subset \mathbb R^3,
    \label{eqa:flow}
\end{equation}
\end{linenomath*} 
where $K(\mathbf x)$ is the hydraulic conductivity of the aquifer $\Omega$, and $h(\mathbf x)$ is the hydraulic head. This PDE is subject to appropriate boundary conditions on the simulation domain boundary $\partial \Omega$. After the flow equation is solved, the average pore velocity $\mathbf u(\mathbf x) = (u_1,u_2,u_3)^\top$ is computed from Darcy's law,
\begin{linenomath*}
\begin{equation} \label{eqa:velo}
    \mathbf{u} = -\frac{K}{\theta} \nabla h,
    \end{equation}
\end{linenomath*}
where $\theta(\mathbf x)$ is the aquifer's porosity. 

Starting at some unknown time $t_0$, a contaminant with volumetric concentration $c_\text{s}$ enters the aquifer through either point-wise or spatially distributed sources $\Omega_\text{s} \subset \Omega$. The contaminant is released for an unknown duration $T$ with unknown intensity $q_\text{s}(\mathbf x, t)$ (volumetric flow rate per unit source volume), such that $q_\text{s}(\mathbf x, t) \neq 0$ for $t_0 \le t \le t_0 + T$. The contaminant is advected by the flow, while undergoing hydrodynamic dispersion and sorption to the solid matrix with rate $R_n$.  Without loss of generality, the spatiotemporal evolution of the contaminant's volumetric concentration $c(\mathbf x,t)$ is described by an advection-dispersion-reaction equation:
\begin{linenomath*}
\begin{equation}\label{eqa:trans_full}
    \frac{\partial \theta c}{\partial t} = \nabla \cdot (\theta \mathbf{D}\nabla c) - \nabla \cdot(\theta \mathbf{u}c) - R_n(c) + q_\text{s} c_\text{s}, \qquad \mathbf x \in \Omega, \quad t > t_0,
\end{equation}
\end{linenomath*}
where the dispersion coefficient $\mathbf{D}$ is a semi-positive second-rank tensor. If the coordinate system is aligned with the mean flow direction, such that $\mathbf u = (u \equiv | \mathbf u |, 0,0 )^\top$, then the components of this tensor are: 
\begin{linenomath*}
\begin{equation}\label{eqa:disper}
      D_{11} = \theta D_\text{m} + \alpha_L u, \quad 
      D_{22} = \theta D_\text{m} + \alpha_T u, \quad
      D_{33} = \theta D_\text{m} + \alpha_C u, \quad
      D_{ij} = \theta D_\text{m} ~~\text{for} ~~i\ne j,
\end{equation}
\end{linenomath*}
where $D_\text{m}$ is the coefficient of molecular diffusion for the contaminant in free water; $\alpha_L$ is the  longitudinal dispersivity; and $\alpha_T$ and $\alpha_C$ are transverse dispersivities in the $x_2$ and $x_3$ directions, respectively.  

The chemical reactions considered represent sorption of the dissolved contaminant onto the solid surface of the porous media. 
Thus, the reaction terms $R_n(c)$ take the form:
\begin{equation}
    R_n(c) = -\rho_b \frac{\partial \tilde{c}}{\partial t}, 
\end{equation}
where $\rho_b$ is the bulk density and $\tilde{c}$ is the concentration sorbed. 

We assume the system to be in local chemical equilibrium, i.e., sorption to be much faster than advection and dispersion. We also assume that sorption does not affect the porosity $\theta$, which remains constant throughout the simulations. With these assumptions,~\eqref{eqa:trans_full} reduces to:
\begin{equation}
    R\theta \frac{\partial c}{\partial t} = \nabla \cdot (\theta \mathbf{D}\nabla c) - \nabla \cdot(\theta \mathbf{u}c) +  q_s c_s,
    \label{eqa:trans}
\end{equation}
wherein $R(c)$ is the dimensionless retardation factor defined as:
\begin{equation}
    R =1 + \frac{\rho_b}{\theta} \frac{\partial \tilde{c}}{\partial c}.
\end{equation}
A sorption isotherm defines the relationship between the sorbed concentration, $\tilde{c}$, and the dissolved concentration, $c$. Among the popular isotherms---linear, Langmuir, and Freundlich---we adopt the latter, for the sake of concreteness. According to the Freundlich isotherm,
\begin{equation}\label{eq:Freundlich}
    \tilde{c} = K_f c^a,
\end{equation}
where $K_f$ is the Freundlich constant, $(L^3M^{-1})^a$; and $a$ is the Freundlich exponent. The units of all relevant transport quantities are summarized in Table~\ref{tab:trans_terms}. \textcolor{blue}{The ``Type'' column describe if the terms are known in the experiments, simulated with given conductivity field and source terms, or estimated as the unknown information.}

\begin{table}[H]
    \centering
    \caption{ Quantities in the transport model~\eqref{eqa:trans} and their units.}
    \begin{tabular}{l l l l}
        Term & Physical quantity & Units & Type\\
        \hline
        $c$ & dissolved concentration & ML$^{-3}$ & simulated \\
        $\theta$ & porosity of the subsurface medium & - & known \\
        $x_i$ & the distance along the respective Cartesian coordinate axis & L & - \\
        $D_{ij}$ & hydrodynamic dispersion coefficient tensor & L$^2$T$^{-1}$ & simulated\\
        $u_i$ & pore water velocity & LT$^{-1}$ & simulated\\
        $q_s$ & volumetric flow rate per volume, sources ($+$) and sinks ($-$) & T$^{-1}$ & known\\
        $c_s$ & concentration of source or sink flux & ML$^{-3}$ & estimated\\
        $R_n$ & chemical reaction term & ML$^{-3}$T$^{-1}$ & simulated \\
        $\rho_b$ & bulk density of the medium & ML$^{-3}$  & known\\
        $\tilde{c}$ & concentration sorbed & ML$^{-3}$ & simulated\\
        $K_f$ & Freundlich constant & (L$^3$M$^{-1}$)$^a$ & known\\
        $a$ & Freundlich exponent & - & known\\

    \end{tabular}
    \label{tab:trans_terms}
\end{table}

\subsection{Parameters of Interest} 
\label{sec:lastData}

Our goal is to identify the conductivity field $K(\mathbf x)$ and the contaminant source $q_s c_s(\mathbf x,t)$, given the flow and transport models,~\eqref{eqa:flow}--\eqref{eq:Freundlich}, and measurements of contaminant concentration and hydraulic head.  Other parameters in the transport model, such as porosity, reaction term coefficients, etc., are assumed to be known. The contaminant release is temporally discretized into $N_\text{re}$ intervals, with a constant release strength during each time interval. Identification of the source term $q_\text{s} c_\text{s}$ is tantamount to finding the location(s), $\mathbf{S}_\text{l}$, and strength, $\mathbf{S}_ \text{s} \in \mathbb{R}^{N_\text{re}}$, of the contaminant source; with the elements $S_{\text{s}j}$ ($j=1, \dots, N_\text{re}$) of the vector $\mathbf{S}_ \text{s} $ denoting the release strength at the $j$-th time interval.

Measurements of hydraulic head, $\bar{h}_{m} =  \bar{h}(\mathbf{x}_m)$, and solute concentration, $\bar{c}_{mi} =  \bar{c}(\mathbf{x}_m,t_i)$, are  collected at locations $\{\mathbf{x}_m\}_{m = 1}^{M}$ at times $\{t_i\}_{i = 1}^{I}$. In lieu of field observations, we generate these data by corrupting the solution of~\eqref{eqa:flow}--\eqref{eq:Freundlich} obtained for the reference parameter  values by random measurement errors $\epsilon^{c}_{mi}$ and $\epsilon^{h}_{m}$, such that:
\begin{linenomath*}
\begin{align}\label{eq:data_mcmc}
\bar{c}_{m,i}  = c(\mathbf{x}_m,t_i) + \epsilon^{c}_{mi}, \qquad
\bar{h}_{m} =  h(\mathbf{x}_m) + \epsilon^{h}_{m}; \qquad m = 1,\dots,M, \quad i = 1,\dots,I,
\end{align}
\end{linenomath*}
where $c(\mathbf{x}_m,t_i)$ and $h(\mathbf{x}_m) $ are the model predictions.  The zero-mean Gaussian random variables $\epsilon^{c}_{mi}$ have covariance $\mathbb{E}[\epsilon^{c}_{mi}\epsilon^{c}_{nj}] = \delta_{ij} R^{c}_{mn}$, where $\mathbb{E}[\cdot]$ denotes the ensemble mean; $\delta_{ij}$ is the Kronecker delta function; and $R^{c}_{mn}$ with $m, n \in [1,M]$ are components of the $M\times M$ spatial covariance matrix $\mathbf{R}^{c}$ of measurements errors. To be specific, we set  $\mathbf{R}^{c} = \sigma_c^2 \mathbf I$, where $\sigma_c$ is the standard deviation of the measurement errors, and $\mathbf I$ is the $(M \times M)$ identity matrix. The hydraulic head measurement errors $\epsilon^{h}_{m}$ are zero-mean Gaussian random variables with covariance $\mathbb{E}[\epsilon^{h}_{m}\epsilon^{h}_{n}] = R^{h}_{mn}$ with $m,n \in [1,M]$. We set $\mathbf{R}^{h} = \sigma_h^2 \mathbf I$, where $\sigma_h$ is the standard deviation of the measurement errors.

The error model in~\eqref{eq:data_mcmc} assumes the flow and transport models~\eqref{eqa:flow}--\eqref{eq:Freundlich} to be exact and the measurements errors to be unbiased and uncorrelated in time but not in space. The groundwater flow equation is solved with MODFLOW~\cite{harbaugh2005modflow}, and the solute transport equation with  MT3DMS~\cite{zheng1999mt3dms,bedekar2016mt3d}. The latter employs a standard finite-difference method with upstream or central-in-space weighting. We use \texttt{Flopy}~\cite{bakker2016scripting}, a \texttt{Python} implementation of these two packages.

%%%%%%%%%%%%%%%%%%%%%%%% III Methodology %%%%%%%%%%%%%%%%%%%%%%%
\section{Methodology}\label{sec:methods}
Below we describe the three elements of our inversion framework: ensemble smoother with multiple data assimilation (ESMDA), convolutional adversarial autoencoder (CAAE) parameterization of the conductivity field, and a Dense encoder-decoder (DenseED) neural network surrogate of the forward model.

\subsection{Ensemble Smoother with Multiple Data Assimilation (ESMDA)}\label{sec:ESMDA}

Upon a spatiotemporal discretization, the uncertain (random) input parameters in~\eqref{eqa:flow}--\eqref{eqa:disper} are rearranged into a vector $\mathbf m$ of length $N_m$; these inputs include the discretized source term $(\mathbf S_\text{l}$, $\mathbf S_\text{s})$ and hydraulic conductivity $K(\mathbf x)$ in all discretized cells (in the applications in this study, $K(\mathbf{x})$ is parameterized with a low dimensional variable, illustrated in detail in Section~\ref{sec:CAAE}). Similarly, we arrange the random measurements $\bar{c}_{m,i}$ and $\bar{h}_{m}$ into a vector $\mathbf d$ of length $N_{d} = M(I+1)$, and the random measurement noise $\epsilon^{c}_{mi}$ and $\epsilon^{h}_{m}$ into a vector $\boldsymbol{\varepsilon}$ of the same length. Then, the error model~\eqref{eq:data_mcmc} takes the vector form,
\begin{linenomath*}
\begin{equation}\label{eq:data-vec}
\mathbf d = \mathbf g(\mathbf m) + \boldsymbol{\varepsilon},
\end{equation}
\end{linenomath*}
where $\mathbf g(\cdot)$ is the vector, of length $N_d$, of the correspondingly arranged stochastic model predictions $c(\mathbf{x}_m,t_i)$ and $h(\mathbf{x}_m)$ predicated on the model inputs $\mathbf m$. Let $\pi (\mathbf{m})$ denote a prior PDF of the inputs $\mathbf m$, which encapsulates the knowledge about the aquifer's properties and contaminant source before any measurements are assimilated. Our goal is to improve this prior by assimilating the measurements $\mathbf d$, i.e., to compute the posterior PDF of the model parameters, $\pi (\mathbf m| \mathbf d)$. This task is accomplished via Bayes' rule,
\begin{equation}\label{eqa:likelihood}
\pi (\mathbf m| \mathbf d) = \frac{\pi (\mathbf{m}) \pi(\mathbf d | \mathbf m)} {\pi (\mathbf d)}, 
\qquad 
\pi (\mathbf d) = \int \pi (\mathbf{m}) \pi(\mathbf d | \mathbf m) \text d \mathbf m,
\end{equation}

where $\pi (\mathbf d| \mathbf m)$ is the likelihood function; and $\pi (\mathbf{d})$, is the ``evidence'' that serves as a normalizing constant so that $\pi (\mathbf{m}) \pi(\mathbf d | \mathbf m)$ integrates to $1$.

To compute~\eqref{eqa:likelihood}, we use ESMDA~\cite{Emerick2013}, which is an ensemble updating method similar to ensemble smoother (ES)~\cite{van1996data} or ensemble Kalman filter (EnKF)~\cite{evensen1994sequential,evensen2003ensemble}. To place ESMDA in the proper perspective, we briefly describe ES. The method is initiated by drawing $N_\text{e}$ samples $\mathbf{M}^f = \{\mathbf{m}^f_1, \dots, \mathbf{m}^f_{N_\text{e}}]$ from the prior PDF $\pi (\mathbf{m})$. These models are then linearly updated as
\begin{linenomath*}
\begin{equation}\label{eqa:ES}
\mathbf{m}_j^a = \mathbf{m}_j^f + \mathbf{C}_{\mathbf{MD}}^f ( \mathbf{C}_{\mathbf{DD}}^f + \mathbf{C_D})^{-1} [ \mathbf{d}_{uc,j} - g(\mathbf{m}_j^f)], \qquad j=1, \ldots, N_\text{e},
\end{equation}
\end{linenomath*}
forming $\mathbf{M}^a = [\mathbf{m}^a_1, \ldots, \mathbf{m}^a_{N_\text{e}}]$, the updated ensemble conditioned on the measurements $\mathbf{d}$. Here,
$\mathbf{C_D} \in \mathbb{R}^{N_d \times N_d}$ is the covariance matrix of the measurement errors $\mathbf{\varepsilon}$; we define an ensemble of perturbed measurements: $\{ \mathbf{d}_{uc,j}\}_{j=1}^{N_e}$, which are obtained by sampling from the Gaussian distribution: $\mathbf{d}_{uc,j} \sim \mathcal{N}(\mathbf{d}, \mathbf{C_D})$; $\mathbf{C}^f_{\mathbf{DD}} \in \mathbb{R}^{N_d \times N_d}$ is the auto-covariance matrix of the model predictions $\mathbf{D}^f = \mathbf{D}^f = [g(\mathbf{m}_1^f), \ldots , g(\mathbf{m}_{N_e}^f)]$; and $\mathbf{C}^{f}_{\mathbf{MD}} \in \mathbb{R}^{N_m \times N_d}$ is the cross-covariance matrix between $\mathbf{M}^{f}$ and $\mathbf{D}^{f}$. During the update, all the data $\mathbf{d}$ are used once, simultaneously. This global update may cause an unacceptably large mismatch between the model response and the measurements, which precipitated the development of an iterative ES with smaller-scale updates.

While ES performs a single large Gauss-Newton correction to the ensemble $\mathbf M^f$, ESMDA makes a smaller correction during each update and deploys the inflated covariance matrix $\mathbf{C_{D}}$ to damp the changes in the ensemble at early iterations~\cite{gao2004improved, wu1999conditioning}. (In the linear Gaussian case, ESMDA  and ES yield identical results.) We use the following algorithm to implement ESMDA.
\begin{itemize}
    \item Set the number of data assimilation iterations, $N_a$, and the corresponding inflation coefficients $\alpha_i$ $i=1, \dots, N_a$. The requirement $\sum_{i=1}^{N_a}\alpha_i=1$ guarantees consistency with ES in the linear Gaussian case; it acts as a constraint for the ES-MDA method in general. Generate the initial ensemble $\mathbf{m}_j^1$ ($j=1, \dots, N_\text{e})$ from the prior PDF $\pi (\mathbf{m})$.
    \item Repeat the following steps for $i=1,\dots,N_a$:
    \begin{enumerate}
        \item Run the forward simulation for each member $\mathbf{m}_j^i$ ($j=1, \dots, N_\text{e}$) from the parameter ensemble $\mathbf M^f$ to obtain the corresponding model predictions (and in the synthetic case, observations) $g(\mathbf{m}_j^i)$.
        \item Perturb the measurements with inflated measurement noise: $\mathbf{d}_{uc,j}^i \sim \mathcal{N}(\mathbf{d}, \alpha_i \mathbf{C_D})$.
        \item Compute the cross covariance matrix $\mathbf{C}^i_{\mathbf{MD}}$ and the auto-covariance matrix of the predicted data $\mathbf{C}^i_{\mathbf{DD}}$.
        \item Update the ensemble as in~\eqref{eqa:ES}, but with $\mathbf{C_D}$ replaced by $\alpha_i \mathbf{C_D}$:
        \begin{linenomath*}
        \begin{equation}\label{eqa:ESMDA}
        \mathbf{m}_j^{i+1} = \mathbf{m}_j^{i} + \mathbf{C}^{i}_{\mathbf{MD}} ( \mathbf{C}^{i}_{\mathbf{DD}} + \alpha_i \mathbf{C_D})^{-1} [ \mathbf{d}_{uc,j}^{i} - g(\mathbf{m}_j^i)], \qquad j=1, \ldots, N_\text{e}.
        \end{equation}
        \end{linenomath*}
    \end{enumerate}    
\end{itemize}
The inverse, $\mathbf{C}_i^{-1}$, of the matrix $\mathbf{C}_i = \mathbf{C}^{i}_{\mathbf{DD}} + \alpha_i \mathbf{C_D}$ is approximated by its pseudo-inverse using a truncated singular value decomposition (TSVD).

\subsection{CAAE Parameterization of Conductivity Field}
\label{sec:CAAE}

Let the matrix $\mathbf{k} \in \mathbb R^{W\times H \times D}$ denote the log-conductivity field $\ln K(\mathbf x)$ defined on a three-dimensional numerical grid, which consists of $W$, $H$ and $D$ elements in the three spatial directions. We use CAAE to parameterize the high-dimensional $\mathbf{k}$ with a low-dimensional latent variable $\mathbf{z}$. The CAAE consists of two components, a GAN and an autoencoder (AE). 

The GAN~\cite{goodfellow2014generative} is a DNN strategy for generating data from complex distributions without having to actually acquire the full PDF. This strategy comprises two networks: a generator $\mathcal{G}(\cdot)$ that generates samples similar to  $\mathbf{k}$; and a discriminator $\mathcal{D}(\cdot)$ that is trained to distinguish between the generated samples and the real data samples. By ``playing an adversarial game'', the discriminator $\mathcal{D}(\cdot)$ improves its ability to catch flaws in the generated samples, and the generator $\mathcal{G}(\cdot)$ improves its capacity to generate realistic samples that try to trick the discriminator. 

The AE learns a low-dimensional representation $\mathbf{z}$ of the data $\mathbf{k}$, and then generates a reconstruction $\mathbf{\hat{k}}$ from $\mathbf{z}$ that closely matches the original data $\mathbf{k}$. The encoded latent variable $\mathbf{z}$ is constructed to follow a PDF $\pi (\mathbf z)$ that is easy to sample from, e.g., a standard normal PDF $\mathcal{N}(\mathbf{0, I})$. A variational autoencoder (VAE)~\cite{kingma2013auto} forces the empirical PDF of $\mathbf z$ computed from the samples of $\mathbf k$, $\pi (\mathbf z | \mathbf k)$, to be close to the target PDF $\pi (\mathbf z)$ by adding the Kullback-Leibler divergence $\mathrm{KL}[\pi (\mathbf z | \mathbf k) \| \pi (\mathbf z)]$ between the empirical and target PDFs to the total loss function: 

\begin{linenomath*}
\begin{equation}\label{eq:VAE-loss}
         \mathcal{L}_\text{VAE} = \mathcal{L}_{\mathrm{rec}}(\mathbf{k},\mathbf{\hat{k}}) +\mathrm{KL}[\pi (\mathbf z | \mathbf k) \| \pi (\mathbf z)]
        \big],
\end{equation}
\end{linenomath*}
where $\mathcal{L}_{\mathrm{rec}}(\mathbf{k},\mathbf{\hat{k}})$ is the discrepancy between the data $\mathbf{k}$ and their reconstruction $\mathbf{\hat{k}}$. Choices of this discrepancy function include $L_1$ or $L_2$ norm. We use the former to define    the average reconstruction error $\mathcal{L}_{\mathrm{rec}}$ over $N$ training samples,
\begin{linenomath*}
\begin{equation}
    \mathcal{L}_{\text{rec}}=\frac{1}{N}\sum_{i=1}^{N}||\mathbf{k}_i-\hat{\mathbf k}_i||_1.
\end{equation}
\end{linenomath*}

The CAAE differs from the VAE in the way it minimizes the discrepancy between the empirical PDF $\pi (\mathbf z | \mathbf k)$ and the target PDF $\pi (\mathbf z)$ of the latent random variable $\mathbf z$. Instead of minimizing the KL divergence $\mathrm{KL}[\pi (\mathbf z | \mathbf k) \| \pi (\mathbf z)]$, the adversarial autoencoder (AAE) employs an adversarial training procedure to minimize this discrepancy. 
The training of the encoder $\mathcal{G}(\cdot)$, decoder $\text{De}(\cdot)$, and the discriminator $\mathcal{D}(\cdot)$ is divided into the reconstruction phase and the regularization phase~\cite{Makhzani2016}. Parameters in the encoder and decoder are updated by minimizing the loss function:
\begin{equation}\label{eq:G-loss}
    \mathcal{L}_{\mathrm{ED}}=\mathcal{L}_{\text{rec}}+w\mathcal{L}_\mathcal{G}.
\end{equation}
We use $\mathcal{L}_\mathcal{G}$ to quantify the decoder's ability to trick the discriminator, 
\begin{linenomath*}
\begin{equation}
    \mathcal{L}_\mathcal{G}=-\frac{1}{N}\sum_{i=1}^{N}\ln\big\{\mathcal{D}[\mathcal{G}(\mathbf{k}_i)]\big\}.
\end{equation}
\end{linenomath*}
The weight factor $w$ in~\eqref{eq:G-loss} is used to assign relative importance to these two sources of error. In the simulations reported below, we set $w=0.01$.

After the encoder and decoder are updated in the first training phase, the discriminator $\mathcal{D}(\cdot)$ is trained in the second phase to minimize the loss function:
\begin{linenomath*}
\begin{equation}\label{eq:D-loss}
    \mathcal{L}_\mathcal{D}=-\frac{1}{N}\sum_{i=1}^{N}\Big\{\ln\big[\mathcal{D}(\mathbf{z}_i)\big]+\ln{\big[1-\mathcal{D}[\mathcal{G}(\mathbf{k}_i)\big]}\Big\}.
\end{equation}
\end{linenomath*}
By iterating between these two training phases, one obtains the mappings from $\mathbf{k}$ to $\mathbf{z}$ and from $\mathbf{z}$ to $\mathbf{\hat{k}}$, and the decoder reaches its goal of constructing realizations $\mathbf{\hat{k}}_i$ similar to the  data $\mathbf{k}_i$.

The architectures of each network in the CAAE in this study are adopted from~\cite{mo2019integration}, and illustrated with our modified schematics in SI. We applied slight modifications to fit the dimensions and specifics of the problem in this study.

\subsection{DenseED Neural Networks as Forward Model Surrogates}\label{sec:cnn}

ESMDA inversion requires a large number of forward solves of the PDE-based model~\eqref{eqa:flow}--\eqref{eq:Freundlich} for multiple realizations of the parameters $\mathbf m$. To alleviate the cost of each forward run, we replace the PDE-based model with a CNN surrogate. 

Several approaches to constructing an input-output surrogate are collated in Table~\ref{tab:1to1-auto}. We choose an autoregressive   model over a one-to-many   model based on computer-memory considerations: for three-dimensional problems with $I$ time steps, memory allocated for input and output can be prohibitively large; also, the autoregressive scheme reduces the number of DNN parameters needed for the regression task. The autoregressive structure enables us to predict the full images (image-to-image) at each time step. That strategy has a superior generalizability than its image-to-sensors counterparts, which  predict concentration values only at sparse locations where measurements are collected \cite{zhou2021markov}.

The source location ($\mathbf S_{\text{l},t}$) and strength ($\mathbf S_{\text{s},t}$) for the release period $[t,t+\Delta t]$ are assembled into a three-dimensional matrix $\mathbf S(\mathbf{x},t) \in \mathbb R^{W\times H \times D}$. 

\begin{table}[htbp]
\centering
\caption{ Alternative input-output frameworks for construction of a surrogate model. The data are collected at $M$ locations $\mathbf x_m$ ($m=1,\cdots,M$) at $I$ times $t_i$ ($i=1,\cdots,I$). The source location ($\mathbf S_{\text{l},t}$) and strength ($\mathbf S_{\text{s},t}$) for the release period $[t,t+\Delta t]$ are assembled into a three-dimensional matrix $\mathbf S(\mathbf{x},t)$. }
\begin{tabular}{l l l c}
\hline
  Model & Input & Output & Modeling frequency \\
\hline
  PDE model                                 & $ \mathbf{m}$ & $c(\mathbf x,t_i)$, $h(\mathbf{x})$ & 1  \\
  Image-to-image                          & $\mathbf{m}$ & $c(\mathbf x,t_i)$, $h(\mathbf{x})$  & 1 \\
  Image-to-sensors                       & $\mathbf{m}$ & $c(\mathbf x_{m},t_i)$, $h(\mathbf x_m)$   & 1  \\
  Autoregressive i-to-i & $c(\mathbf{x},t), \ln K(\mathbf{x}),  S(\mathbf{x},t)$ & $c(\mathbf{x},t + \Delta t)$, $h(\mathbf{x})$  & $I$   \\
\hline
\end{tabular}
\label{tab:1to1-auto}
\end{table}

An autoregressive surrogate $\mathbf{NN}_\text{auto}$ replaces the \textcolor{blue}{PDE}-based model: 
\begin{align}
\mathbf g : \mathbf m \xrightarrow{\text{PDEs}} \{c(x_{m},t_i), h(x_{m})\}_{m,i = 1}^{M,I} 
\end{align}
with  a CNN that sequentially ($I$ times) predicts the system state at the next time step, 
\begin{linenomath*}
\begin{align}\label{eqa:surro}
\mathbf{\mathbf{NN}_\text{auto}} : c(\mathbf{x},t_i), K(\mathbf{x}), S(\mathbf{x},t_i) \xrightarrow{\text{CNN}} \{c(\mathbf{x},t_{i+1}), h(\mathbf{x})\}, \qquad i = 0,\dots,I-1. 
\end{align}
\end{linenomath*}

\begin{figure}[htbp]
    \centering
    \includegraphics[width=0.8\textwidth]{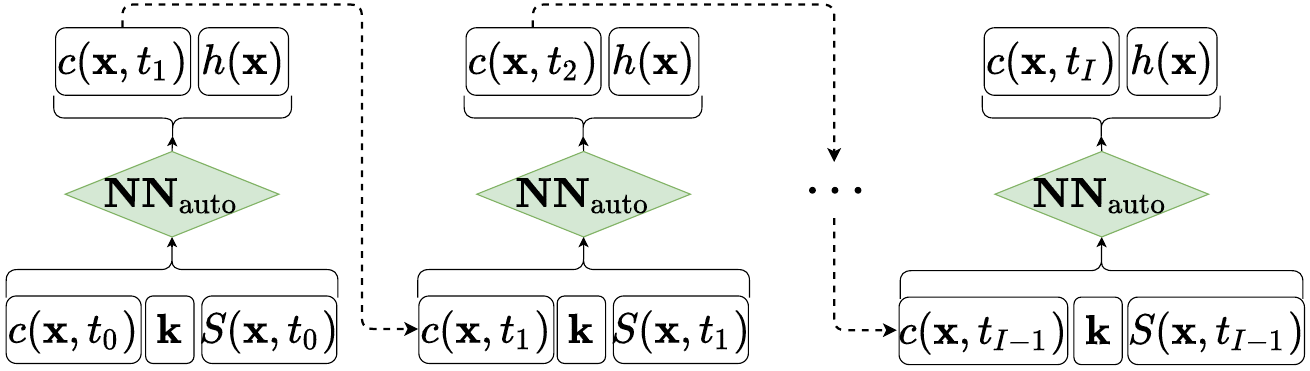}
    \caption{Autoregressive surrogate $\mathbf{\mathbf{NN}_\text{auto}}$ of the PDE-based flow and transport model~\eqref{eqa:flow}--\eqref{eq:Freundlich}. The three input channels correspond to $c(\mathbf{x},t)$, $S(\mathbf{x},t)$, and $\ln K(\mathbf{x})$. The two output channels correspond to $c(\mathbf{x},t+\Delta t)$ and $h(\mathbf{x})$. The concentration values at all time steps ${t_i}_{i=1}^{I}$ are obtained through the iteration prediction with the autoregressive model.}
    \label{fig:nn_auto_diagram}
\end{figure}

If the three-dimensional simulation domain is discretized with a $D\times H\times W$ grid, then the autoregressive CNN surrogate $\mathbf{\mathbf{NN}_\text{auto}}$ performs the following input-to-output mapping:
\begin{linenomath*}
\begin{align}
\mathbf{\mathbf{NN}_\text{auto}}: \mathbb{R}^{n_x\times W\times H \times D}\rightarrow{\mathbb{R}^{n_y\times W\times H \times D}},
\end{align}
\end{linenomath*}
where
$n_x = 3$, denotes the three channels representing the concentration $c(\mathbf{x},t_i)$ and source terms $S(\mathbf{x},t_i)$ at time $t_i$, and the log-conductivity $\ln K(\mathbf{x})$; and $n_y=2$ designates the two output channels representing the concentration $c(\mathbf{x},t_{i+1})$ at time $t_{i+1}$ and the hydraulic head $h(\mathbf{x})$. A representative input-to-output example is shown in  Figure~\ref{fig:nn_auto_diagram}.

We use a three-dimensional DenseED architecture to solve the image-to-image regression task with a coarsen-refine process, with the convolutional operations. The encoder extracts the high-level coarse features of the input maps, while the decoder subsequently refines the coarse features to the full maps~\cite[Fig.~2]{mo2019deep}. We use the $L_1$-norm loss function, the $L_2$-norm weight regularization, and stochastic gradient descent~\cite{bottou2010large} in the CNN training process. A detailed description of this surrogate model and its training procedure can be found in SI and in \cite{mo2019deep}. We have extended their procedure by adding the measurement locations to the loss function. This allows us to penalize the prediction error at these specific locations.

One could improve the accuracy of our surrogate model by deploying an iterative optimization strategy, which would increase its computational cost. We chose not to do so because the convolutional encoder-decoder neural networks proved to be sufficiently accurate surrogates of contaminant transport models when used for inverse problems \cite{mo2019deep,zhou2021markov}. \citeA{mo2019integration} provide a detailed analysis of the number of samples needed to train a three-dimensional surrogate of a problem whose spatial discretization ($6 \times 32 \times 64$) is similar to ours. In the spirit of transfer learning, we adopted their most efficient setting of the surrogate model. This is an empirical choice based on our past studies, and we encourage a thorough study on the performance of a surrogate model in other unexplored applications.

\subsection{CAAE--DenseED--ESMDA Inversion Framework}
\label{sec:inverse_overall}

We combine the CAAE parameterization of the conductivity field with the DenseED CNN surrogate of the forward model to obtain fast and accurate predictions of concentration $c(\mathbf x,t)$ and $h(\mathbf x)$ for a given set of inputs. \textcolor{blue}{We provide the coefficient of determination $R^2$ as the measure of accuracy for forward surrogate.} Then, we utilize ESMDA to identify the unknown parameters, including the conductivity field and the source terms $(\mathbf S_\text{l}, \mathbf S_\text{s})$. \textcolor{blue}{We show the box-plots of the ensembles for the assimilated terms to illustrate the quality and uncertainty of the inversion.} The CAAE parameterization enables one to estimate the discretized log-conductivity field $\mathbf k$ through the latent variable $\mathbf{z}$. Our CAAE--DenseED--ESMDA inversion framework is implemented in the following algorithm.
\begin{enumerate}
    \item Train a CAAE; obtain the decoder $\text{De}(\cdot)$ that maps the low-dimensional latent variable $\mathbf{z}$ back onto the log-conductivity field $\mathbf k$.
    \item Train an autoregressive DenseED CNN $\mathbf{NN}_\text{auto}$ to predict $c(\mathbf x,t)$ and $h(\mathbf x)$ for the input conductivity field and contaminant release history.
    \item Generate the initial input ensemble $\mathbf M^f$ of size $N_\text{e}$, whose elements $\mathbf{m}_j^1$ ($j=1, \dots, N_\text{e}$) are defined as $\mathbf{m}_j^1 = (\mathbf{z}^1_j, \mathbf{S_l}^1_j, \mathbf{S_s}^1_j)^\top$. Here, $\mathbf{z}^1_j \sim \mathcal{N}(\mathbf{0, I_z})$ is the latent variable for the log conductivity field; and $\mathbf{S_l}^1_j$ and $\mathbf{S_s}^1_j$ denote respectively the source location and strength in all release periods, drawn from an appropriate prior distribution.
    \item Perform the ESMDA inversion with $N_a$ data assimilation iterations and the inflation coefficients $\alpha_i$ ($i=1,\dots,N_a$). For $i=1,\dots,N_a$,
        \begin{enumerate}
            \item Obtain the log-conductivity realizations $\mathbf{k}^i_j = \text{De}(\mathbf{z}^i_j)$ with $j=1, \dots, N_\text{e}$;
            \item Form the release configuration $\{\mathbf S_{\text l,j}^i, \mathbf S_{\text s,j}^i \}$ into the input matrix $\mathbf{S}^i_j$, and predict $c(\mathbf x,t)$ and $h(\mathbf x)$ at the measurement times and locations, $\mathbf{NN}_\text{auto}(\mathbf m^i_j)$ for all $j$;
            \item Update the ensemble $\mathbf{m}_j^{i}$ via ESMDA with $\alpha_i$ to obtain $\mathbf{m}_j^{i+1}$. 
        \end{enumerate}
    \item The end result, $\mathbf{m}_j^{N_a+1}$, serves as the final ensemble from which PDFs of the log conductivity field and the contaminant release parameters are estimated.
\end{enumerate}

\begin{figure}[htbp]
    \centering
    \includegraphics[width=0.9\textwidth]{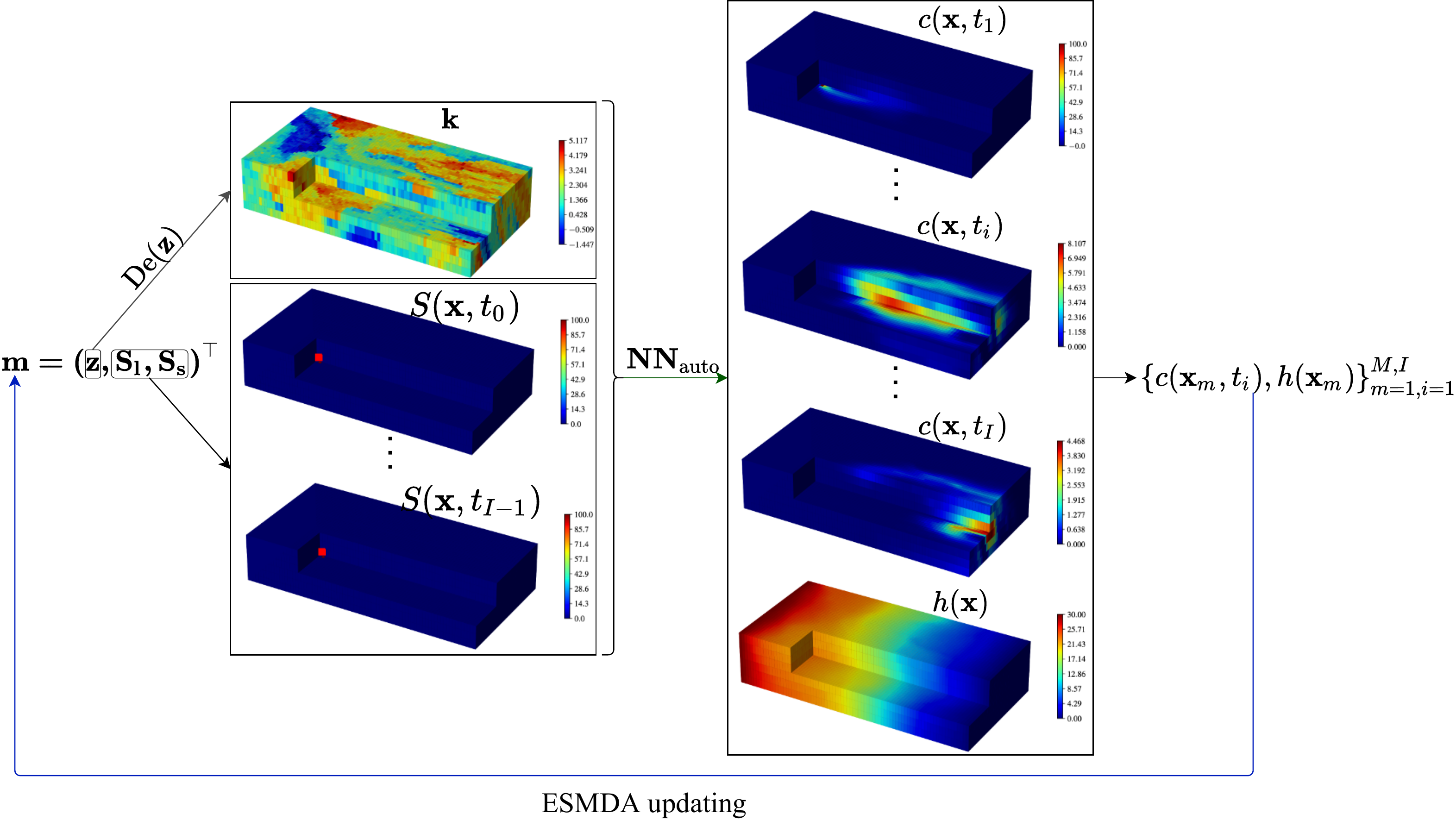}
    \caption{Schematic illustration of the CAAE-DenseED-ESMDA algorithm. $\text{De}(\cdot)$ denotes the decoder that obtains the log-conductivity field from the latent variable $\mathbf{z}$, $\mathbf{NN}_{\rm auto}$ represents the autoregressive surrogate model which predicts the concentration and hydraulic head fields. The observations at the measurement time and locations $c(\mathbf{x}_m, t_i), h(\mathbf{x})_m$ are then used to update the parameter $\mathbf{m}$.}
    \label{fig:ml_schematics}
\end{figure}

This algorithm is illustrated in the schematic  in Figure~\ref{fig:ml_schematics} as well.

%%%%%%%%%%%%%%%%%%%%%% NUMERICAL EXPERIMENTS %%%%%%%%%%%%%%%%%%%%%%%%%%%
\section{Numerical experiments}
\label{sec:results}

\subsection{Experimental Setup}
\label{sec:res_forward}

\begin{figure}[htbp]
    \centering
    \includegraphics[width=0.9\textwidth]{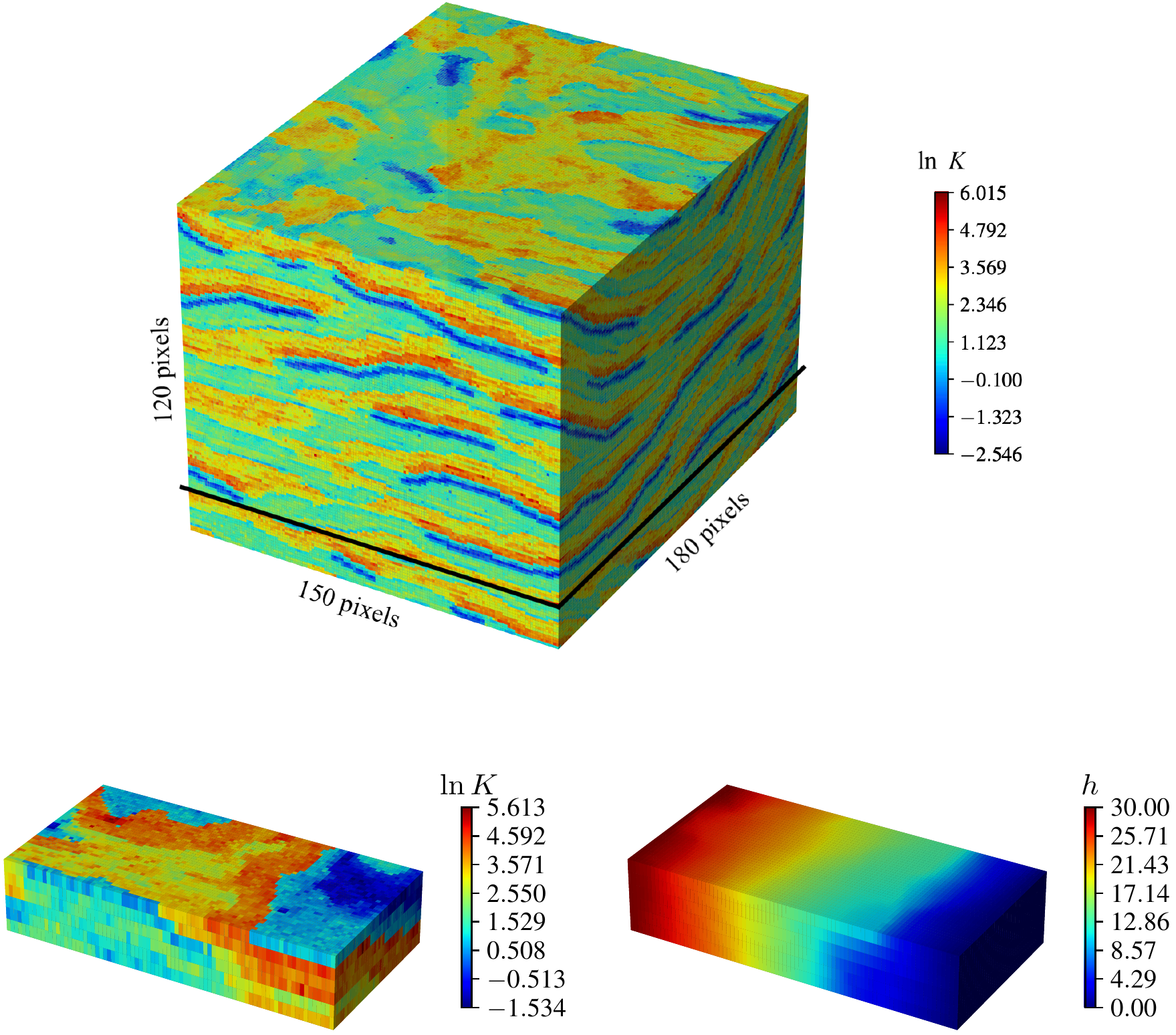}
    \caption{\textbf{Top:} Training image, consisting of $150 \times 180 \times 120$ pixels. Equiprobable realizations of log-conductivity $Y(\mathbf x) = \ln K(\mathbf x)$ are generated by randomly selecting  patches of size $81 \times 41 \times 6$ pixels. The top $150 \times 180 \times 105$ pixels serve as the training set. The bottom $150 \times 180 \times 15$ pixels serve as the testing set. Conductivity $K$ is in m/d. \textbf{Bottom:} Log-conductivity $Y(\mathbf x)$ (left) and the corresponding hydraulic head $h(\mathbf x)$ (right), which serve as ground truth and to generate measurements of $h$ at observation wells. Conductivity $K$ is in m/d and head $h$ in m. }
    \label{fig:logk_training}
\end{figure}

A confined heterogeneous aquifer is described as a rectangular cuboid $\Omega$ of size $2500~\text{m} \times 1250~\text{m} \times 300~\text{m}$; it is discretized with a mesh consisting of $81 \times 41 \times 6$ cells. Groundwater flow is driven by constant heads $h_\text{L} = 30~\text{m}$ and $h_\text{R} = 0~\text{m}$ imposed along the left ($x_1 = 0$) and right ($x_1 = 2500~\text{m}$) facets of the cuboid, respectively; the remaining boundaries are impermeable to flow. The hydraulic conductivity of this aquifer, $K(\mathbf x)$, is unknown (except when generating the ground truth); equiprobable realizations of $Y(\mathbf x) = \ln K(\mathbf x)$ are generated by extracting $81 \times 41 \times 6$ patches from the $150~\text{px} \times 180~\text{px} \times 120~\text{px}$ training image~\cite{mariethoz2011modeling} in Figure~\ref{fig:logk_training},  available at \url{https://github.com/GAIA-UNIL/trainingimages}. One such cropped log-conductivity field $Y(\mathbf x)$ and the corresponding hydraulic head $h(\mathbf x)$, obtained as a solution of the groundwater flow equation~\eqref{eqa:flow}, are shown in Figure~\ref{fig:logk_training}. These fields serve as the ground truth. 

The porosity $\theta$ and bulk density $\rho$ of the soil; dispersivities $\alpha_L$, $\alpha_T$ and $\alpha_C$; and the parameters $K_f$ and $a$ of the Freundlich isotherm are constant \textcolor{blue}{and known}. Values of these transport parameters, which are representative of a sandy alluvial aquifer in Southern California~\cite{liggett2015exploration}, are presented in Table~\ref{tab:trans_para}. The contaminant enters the aquifer via a point source, whose depth is known (the fourth layer from the top of the domain) but the location in the horizontal plane ($S_{\text l}^x$ and $S_{\text l}^y$) is uncertain.  The contaminant release is known to occur during a $20$-year period, but its strength is uncertain. Following the standard practice in groundwater modeling, we divide this time interval into $N_\text{re} = 5$  sub-intervals (``stress periods'' in the MODFLOW/MT3D language) during each of which the release strength ($S_\text{s}$) is constant. In this configuration, the unknown contaminant release history is represented by the vector $\mathbf S = (\mathbf S_\text{l}, \mathbf S_\text{s})$, where $\mathbf S_\text{l} = (S_{\text l}^x, S_{\text l}^y)^\top$ and $\mathbf S_\text{s} = (S_{\text s, 1}, S_{\text s, 2}, S_{\text s, 3}, S_{\text s, 4}, S_{\text s,5})^\top$. The values of $\mathbf S$ used to generate the ground-truth concentrations are reported in Table~\ref{tab:prior_ref_kd3d}. Combined with the discretized version $\mathbf k$ of the uncertain log-conductivity field $Y(\mathbf x)$, this yields  19933 unknowns to be determined from the measurements of solute concentration $c(\mathbf x,t)$ and hydraulic head $h(\mathbf x)$. Expert knowledge about possible location and strength of the contaminant release is encapsulated in the uniform (``uninformative'') prior distributions for $\mathbf S_\text{l}$ and $\mathbf S_\text{s})$, which are shown in Table~\ref{tab:prior_ref_kd3d}.

\begin{table}[htbp]
    \centering
    \caption{Values of the transport parameters for a dissolved contaminant migrating in a generic sandy alluvial aquifer in Southern California~\cite{liggett2015exploration}.}
    \begin{tabular}{llc}
    \hline
        Property & Value & Units\\
    \hline
        $\phi$ & 0.3 & -- \\
        $K_f$ & $0.1$ & $(\text{m}^3 / \text{g})^a$ \\
        $a$ & $0.9$ & -- \\
        $\rho$ & 1587 & kg/m$^3$ \\
        $\alpha_L$ & 35 & m \\
        $\alpha_T / \alpha_L$ & 0.3 & -- \\
        $\alpha_C / \alpha_L$ & 0.3 & -- \\
        $D_\text{m}$ & $10^{-9}$ & m$^2$/d \\
    \hline    
    \end{tabular}
    \label{tab:trans_para}
\end{table}

%%%%%%%%%%%%%%%% Table of prior and true value of the release parameters.
\begin{table}[htbp]
\centering
\caption{Parameters $\mathbf S_\text{l} = (S_{\text l}^x, S_{\text l}^y)^\top$ and $\mathbf S_\text{s} = (S_{\text s, 1}, S_{\text s, 2}, S_{\text s, 3}, S_{\text s, 4}, S_{\text s,5})^\top$ used to represent, respectively, the location and strength of a contaminant release. Reported as ``Truth'' are their (unknown) reference values used to generate ground truth and concentration measurements, and ``Prior'' the intervals on which their uniform priors, $\mathcal U[\cdot,\cdot]$, are defined. The values of $\mathbf S_\text{l}$ are in m, and of $\mathbf S_\text{s}$ in g/m$^3$.}
\begin{tabular}{l c c c c c c c}
\hline
   & $S_{\text l}^x$ & $S_{\text l}^y$ & $S_{\text s, 1}$ & $S_{\text s, 2}$ & $S_{\text s, 3}$ & $S_{\text s, 4}$ & $S_{\text s, 5}$ \\
\hline
  Truth & $291$ & $625$ & $224$ & $174$ & $869$ & $201$ & $741$  \\
  Prior & $[125,625]$ & $[125,1125]$ & $[50,1000]$ & $[50,1000]$ & $[50,1000]$ & $[50,1000]$ & $[50,1000]$ \\
\hline
\end{tabular}
\label{tab:prior_ref_kd3d}
\end{table}

\begin{figure}[htbp]
    \centering
    \includegraphics[width=\textwidth]{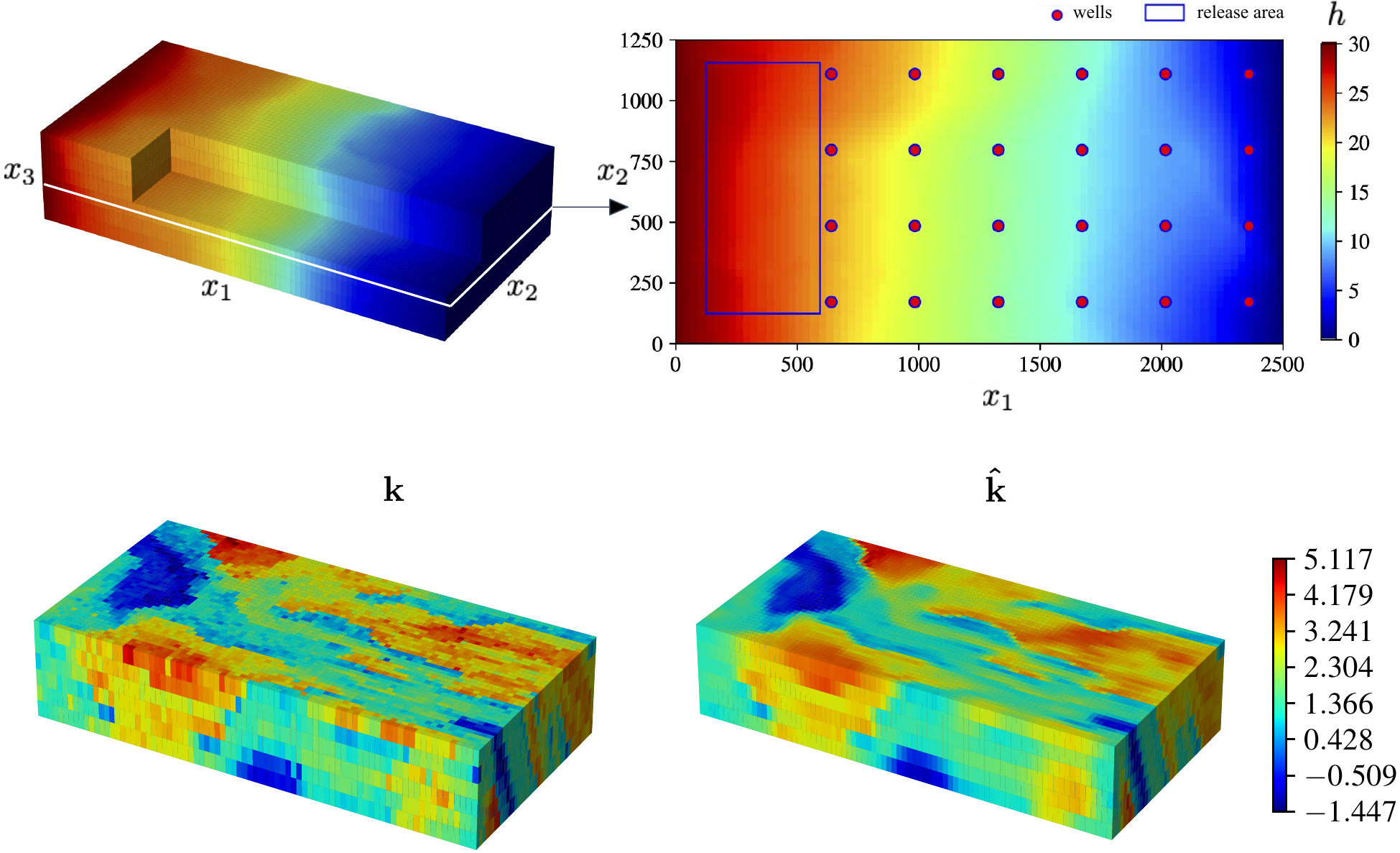}
    \caption{\textbf{Top:} Observational wells (red dots) in which measurements of hydraulic head $h$ and solute concentration $c$ are collected. The well locations are superposed on the ground-truth distribution of hydraulic head in the fourth layer of the MODFLOW model. The blue box represents a region of possible contaminant release from a point source that is known to be located in the fourth model-layer, $x_1,x_2,x_3$ in m. \textbf{Bottom:} A representative realization of $\ln K(\mathbf x) \longmapsto \mathbf k$ from the test set (left) and its reconstruction (right) via the CAAE encoder, $\mathbf{z} = \mathcal{G}(\mathbf k)$, and decoder, $\hat{\mathbf k} = \text{De}(\mathbf z)$.  
     }
    \label{fig:release_candi}
\end{figure}

\textcolor{blue}{The measurements of hydraulic head and contaminant concentration that will be used for data assimilation} are collected at observations wells, whose completion allows one to collect water samples in each of the model's six vertical layers. We consider the observation wells whose locations are depicted in Figure~\ref{fig:release_candi}. 
During the simulated time horizon of $40$ years, the contaminant concentration is sampled at $I = 10$ time intervals of four years each, and the hydraulic head is measured once since the flow is at steady-state. The data at all space-time locations are generated by adding zero-mean Gaussian measurement error with standard deviations $\sigma_c=\sigma_h=0.5$, to the solution $\mathbf{g(m)}$ of the flow and transport model~\eqref{eqa:flow}--\eqref{eq:Freundlich} with the input parameter values identified as ``ground truth'' above, corresponding to $2.5\%$ of the maximum value of the concentration ($\sim 20$ g/m$^3$), and $\sim 1.7\%$ of the maximum value of the hydraulic head ($\sim 30$ m) on all sensor locations. \textcolor{blue}{We summarized all the constants and dimensions of matrices in Table~\ref{tab:const_discr} in Section~\ref{sec:append_c}. We assimilate the data only once for each realization, i.e. we gather all measurements at all $I$ times, and use all these measurements to update the unknown parameters at once.}

\subsection{CAAE Training for Conductivity Parameterization}

We train a CAAE DNN to parameterize the discretized log-conductivity field $\mathbf k \in \mathbb R^{81 \times 41 \times 6}$. The end goal is an encoder $\mathcal G(\mathbf k)$ that maps an input field $\mathbf k$ onto a low-dimensional latent variable $\mathbf z \in \mathbb R^{2\times 2 \times 11 \times 21}$ with standard-Gaussian prior $\mathcal{N}(\mathbf{0,I})$, and a decoder $\text{De}(\cdot)$  that reconstructs $\mathbf k$ from this latent variable. The training is done on 23000 realizations of $\mathbf k$, obtained as randomly selected  ($81 \times 41 \times 6$) patches from the top $150~\text{px} \times 180~\text{px} \times 105~\text{px}$ part of the large training image in Figure~\ref{fig:logk_training}. Additional $2200$ images cropped from the bottom $150~\text{px} \times 180~\text{px} \times 15~\text{px}$ part of the training image serve as the testing set. The latent variable $\mathbf{z}$ has  $2 \cdot 2 \cdot 11 \cdot 21 = 924$ elements. With $50$-epochs training and the learning rate of $2 \cdot 10^{-4}$, the Adam optimizer is used to obtain the DNN parameters and, thus, build $\mathcal G(\mathbf k)$ and $\text{De}(\mathbf z)$. The details of the CAAE architecture are explained in SI, and the dimensions of the internal layers outputs are listed there in Table~S1.

A representative realization of $\ln K(\mathbf x) \longmapsto \mathbf k$ from the test set and its reconstruction via the decoder, $\hat{\mathbf k} = \text{De}(\mathbf z)$, are shown in Figure~\ref{fig:release_candi}. After the training is complete, the mean absolute error $\|\mathbf k - \hat{\mathbf k}\|_1$, averaged over all the elements of the numerical mesh and over the 2200 members of the testing data set, is $0.2637$. The reconstructed log-conductivity field $\hat{\mathbf k}$ captures the main structural features of its original counterpart $\mathbf k$. Some loss of information is unavoidable in reduced-order modeling but, overall, the performance of this autoencoder is adequate to achieve accurate inverse modeling results, as we show in Section~\ref{sec:results_inv} below. In fact, high accuracy of this autoencoder might have a negative impact on the inversion results \cite{lopez2021deep}, which also provides an analysis on the trade off between the inversion quality and the generative autoencoder accuracy.

In addition to the mean absolute error of CAAE, we also run the PDE-based forward model with the reconstructed $\hat{\mathbf k}$, and computed the coefficient of determination $R^2$ for the concentration and hydraulic head fields. This essential experiment indicates the impact of the error in the representation of the heterogeneity on the prediction of the concentration field. The steps for this test are as follows.
\begin{itemize}
    \item Obtain $N_\text{test}$ forward model inputs:  $\mathbf{m}_i = (\mathbf{k}_i, \mathbf{S_l}_i, \mathbf{S_s}_i)^\top, i=1,\dots,N_\text{test}$, where $\mathbf{k}_i$ is the conductivity field sampled from the testing set in Figure~\ref{fig:logk_training}.
    \item Obtain the concentration and hydraulic head predictions with the PDE-based forward model $\mathbf g$: $\mathbf{y}_i = \mathbf{g(m}_i), i=1,\dots,N_\text{test}$. 
    \item Obtain the reconstructed conductivity fields $\{\hat{\mathbf{k}}_i\}_{i=1}^{N_\text{test}}$ by applying the encoder and decoder obtained from the CAAE training: $\hat{\mathbf{k}}_i = \text{De}(\mathcal{G}(\mathbf k))$.
    \item Substitute the conductivity field with the reconstructed ones $\hat{\mathbf k}$, and run the PDE-based forward model again for the prediction: $\hat{ \mathbf y}_i = \mathbf{g(\hat{m}}_i), \mathbf{\hat{m}}_i = (\mathbf{\hat{k}}_i, \mathbf{S_l}_i, \mathbf{S_s}_i)^\top$.
    \item Compute the coefficient of determination, \[
    R^2=1- \frac{\sum^{N_\text{test}}_{i} ||\mathbf{y}_i-\hat{\mathbf y}_i||^2 }{\sum^{N_\text{test}}_{i} ||\mathbf{y}_i-\bar{\mathbf y}||^2}, \qquad \bar{\mathbf y} = \frac{1}{N_\text{test}} \sum_{i=1}^{N_\text{test}}\mathbf{y}_i.\]
\end{itemize}

The evaluated $R^2$ on the $150$ set of forward simulations is $0.9623$.

\subsection{DenseED Surrogate Model}

As mentioned in Section~\ref{sec:methods}, although only model predictions at the well locations are necessary for the inversion, a DNN that predicts $c(\mathbf x, t_i)$ and $h(\mathbf{x})$ at all points $\mathbf x$ of the simulation domain has better generalization properties. We train our CNN on $N = 800$ Monte Carlo realizations of the PDE-based model~\eqref{eqa:flow}--\eqref{eq:Freundlich} with corresponding realizations of the input parameters $\mathbf m$ (the discretized log-conductivity $\mathbf k$ and contaminant release history $\mathbf S$). Another set of $N_\text{test} = 150$ realizations are retained for testing. These $950$ realizations form $950\times 10$ autoregressive input-output pairs. The CNN contains three dense blocks with $N_l = 3$, $6$, and $3$ internal layers, has the growth rate of $R_g = 48$ and $N_\text{in} = 48$  initial features; it was trained for 200 epochs with the learning rate of $5 \cdot 10^{-3}$. We use the $L_1$-norm loss function and the $L_2$-norm weight regularization, apply stochastic gradient descent~\cite{bottou2010large} in the parameter estimation process, and add $5$ times the $L_1$-norm loss at the source pixel and its surrounding pixels, $5$ times the $L_1$-norm loss at the well locations to the total loss to penalize the prediction error at the source locations and the observation wells. The CNN's output is the hydraulic head $h(\mathbf x)$ and the solute concentration $c(\mathbf x, t_{i})$ at the next time step $t_{i}$. The details of the architecture are explained in SI, and the dimensions of the internal layers outputs are listed there in Table~S2.

\begin{figure}[htbp]
    \centering
    \includegraphics[width=\textwidth]{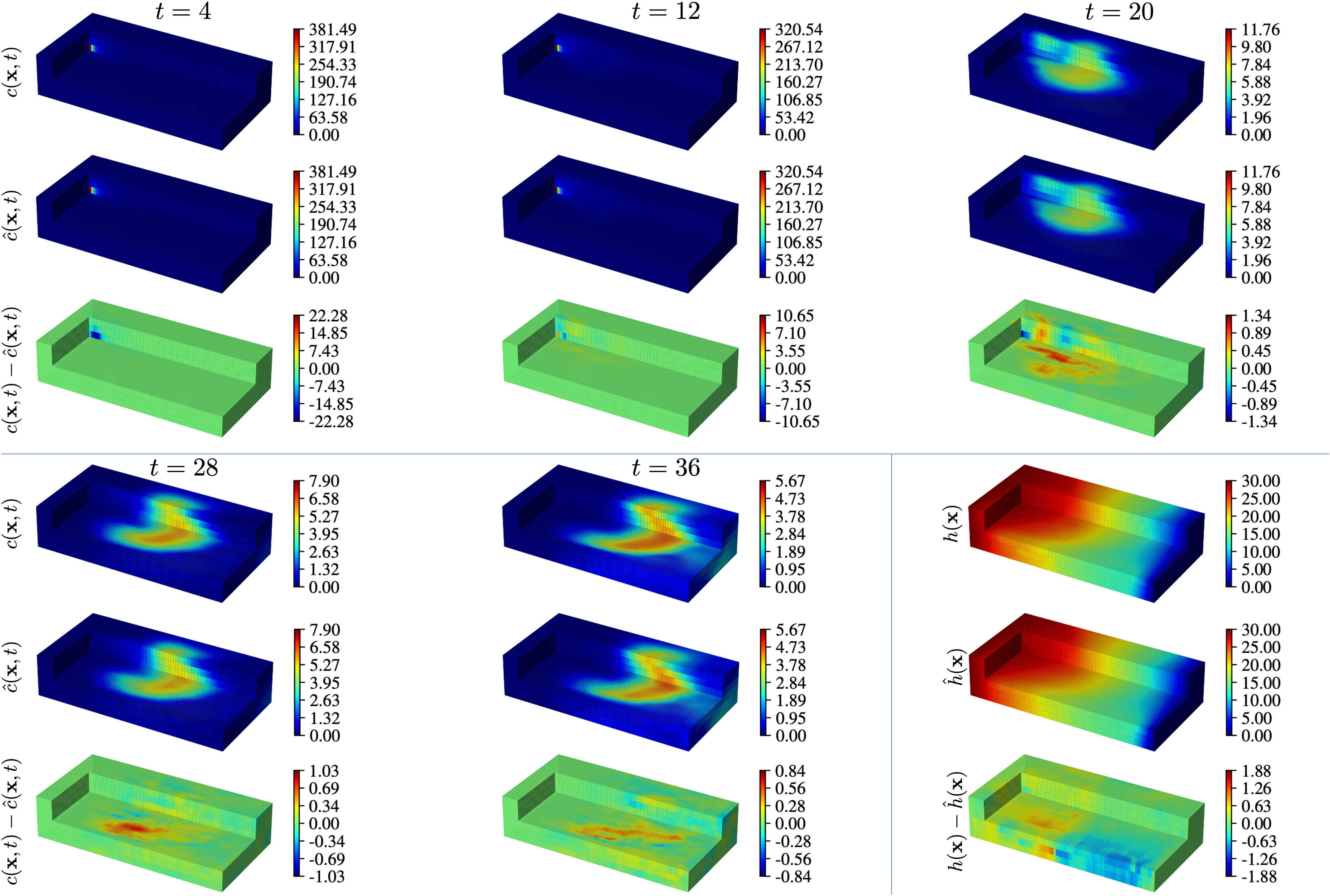}
    \caption{Predictions of the solute concentration obtained with the PDE-based model, $c(\mathbf{x},t)$, and its DenseED CNN surrogate, $\hat{c}(\mathbf{x},t)$, times $t = 4, 12, 20, 28, 36$, $t$ in year. Also shown are the corresponding predictions of the hydraulic head, $h(\mathbf{x})$ and $\hat{h}(\mathbf{x})$; and the difference between these two types $c(\mathbf{x},t)-\hat{c}(\mathbf{x},t), h(\mathbf{x})-\hat{h}(\mathbf{x}) $ of prediction. }
    \label{fig:surro_nn_3dkd}
\end{figure}

Figure~\ref{fig:surro_nn_3dkd} exhibits temporal snapshots of the solute concentrations alternatively predicted with the PDE-based model solved with MODFLOW and MT3DMS simulators, $c(\mathbf x, t_i)$, and the CNN surrogate, $\hat c(\mathbf x,t_i)$, for a given realization of the log-conductivity field and the contaminant release configuration (both drawn from the test set). Also presented are the hydraulic head maps predicted by the autoregressive model, $\hat h(\mathbf x)$, and the PDE-based model solved with MODFLOW simulator, $h(\mathbf x)$.  The accuracy of our CNN surrogate is quantified by the total root mean square error, $(\| c(\mathbf x, t) - \hat c(\mathbf x, t) \|_2 + \| h(\mathbf x) - \hat h(\mathbf x) \|_2) / 2$. It falls to $0.853$ at the end of the training process. The coefficient of determination is $R^2 = 0.79$. It is worthwhile emphasizing here that the $N_\text{NMC} = 800$ Monte Carlo realizations used to train the CNN surrogate are but a small fraction of the forward runs required by ESMDA inversion framework. One could achieve more accurate predictions for three-dimensional problems by either deploying a more complex DNN architecture~\cite{wen2021towards, mo2019integration} or using much larger $N_\text{NMC}$ or both. However, similar to the CAAE training, we focus on the development of efficient methodologies for three-dimensional inverse modeling that accommodate the trade-off between the accuracy and computational feasibility. 

\subsection{ESMDA Inversion}
\label{sec:results_inv}

We demonstrate the use of the CAAE parameterization and the DenseED CNN surrogate of the PDE-based forward model to accelerate the ESMDA inversion. The combination of these three techniques constitutes our CAAE-DenseED-ESMDA framework to approximate the joint posterior PDF of the uncertain model parameters $\mathbf m$ consistent with both model predictions and field observations. In the simulations reported below, we select $N_a = 10$ inflation factors in~\eqref{eqa:ESMDA} and set their values to $\alpha_i = 10$ for $i = 1,\dots,N_a$, and perform ESMDA with 10 iterations. To ascertain the impact of  the DenseED CNN surrogate on the inversion accuracy, we also run CAAE-ESMDA with the PDE-based forward model implemented in MODFLOW and MT3DMS. The ensemble size for ESMDA for both CAAE-ESMDA and CAAE-DenseED-ESMDA are set to $N_\text{e} = 960$. 

The measurements are taken at $24$ wells that are completed in all $6$ layers of the model, yielding $24 \cdot 6 = 144$ measurements of the solute concentration and hydraulic head at each observation time, the hydraulic head is only measured once, resulting in $144 \cdot(10+1) = 1584$ measurements in $40$ years of the modeling time. These measurements are generated with the hydraulic conductivity field $\mathbf{k}$ shown in Figure~\ref{fig:release_candi}.

\begin{figure}[htbp]
    \centering
    \includegraphics[width=\textwidth]{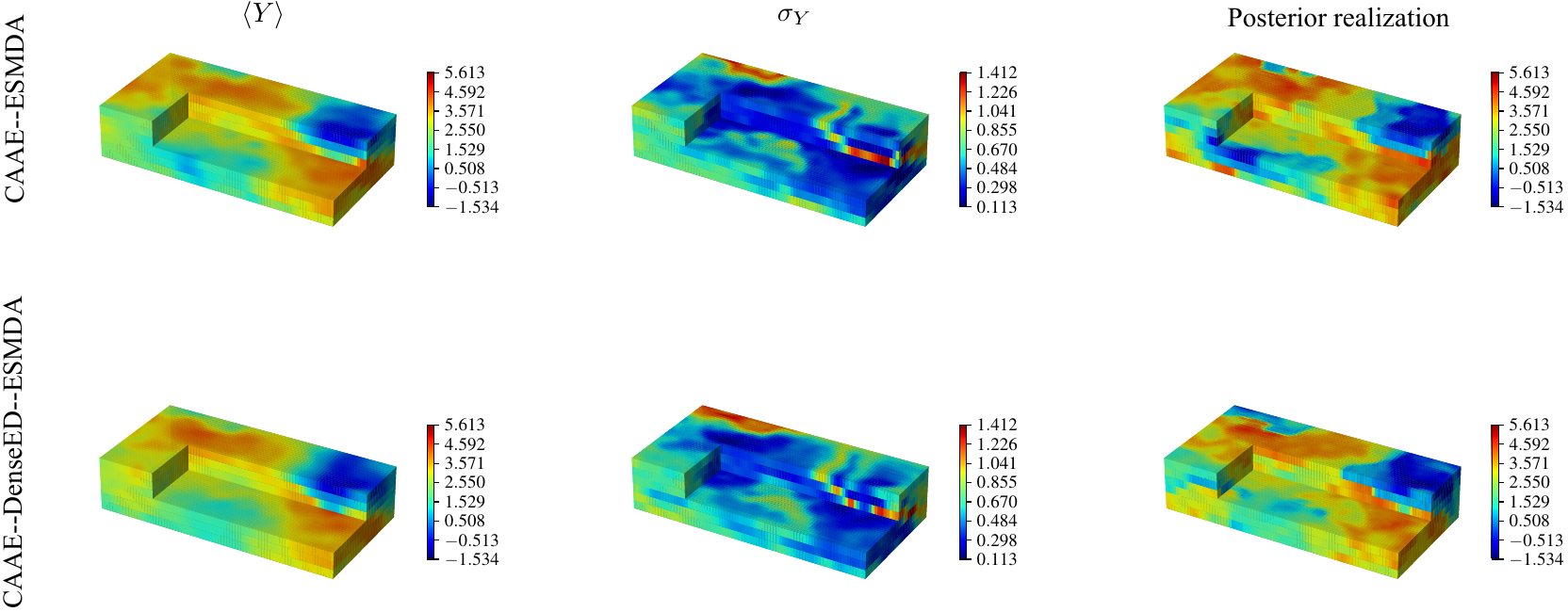}
    \caption{Posterior mean ($\langle Y \rangle$, left column) and standard deviation ($\sigma_{Y}$, middle column) of the log-conductivity field $Y(\mathbf x)$ obtained upon assimilation of concentration and head measurements from the dense observation network. These statistics are obtained via our inversion algorithm CAAE-ESMDA that relies on either the PDE-based forward model (top row) or its DenseED CNN surrogate (CAAE-DenseED-ESMDA, bottom row). Also shown are representative realizations from the resulting posterior ensemble (right column). 
    }
    \label{fig:logk_esmda}
\end{figure}

\begin{figure}[htbp]
    \centering
    \includegraphics[width=0.5\textwidth]{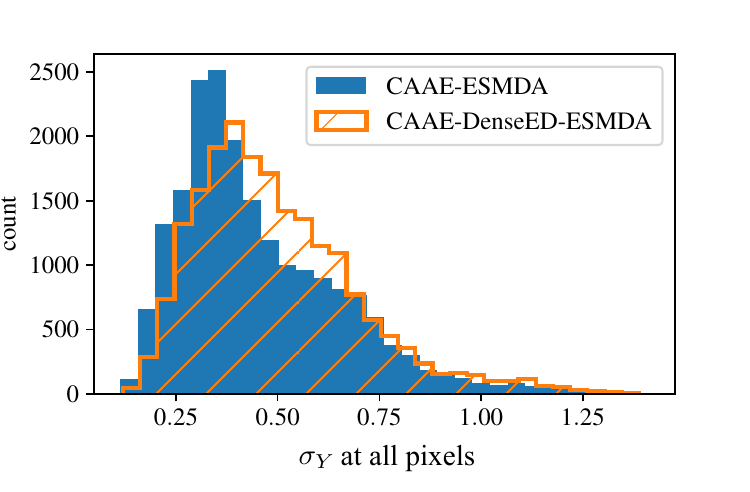}
    \caption{\textcolor{blue}{Histogram of the standard deviation $\sigma_{Y}$ on all pixels for CAAE-ESMDA and CAAE-DenseED-ESMDA.}}
    \label{fig:hist_std_logk}
\end{figure}

Figure~\ref{fig:logk_esmda} exhibits posterior statistics (mean $\langle Y \rangle$ and standard deviation $\sigma_Y$) of the log-conductivity $Y(\mathbf x)$, obtained after the assimilation of all $1584$ measurements via either CAEE-ESMDA or CAEE-DenseED-ESMDA. In both scenarios, the posterior ensemble mean $\langle Y \rangle$, reconstructed from the latent variable $\mathbf{z}$, correctly identifies the low-conductivity region in the right top region of the three-dimensional domain and the high-conductivity regions elsewhere. As expected, the mean log-conductivity fields, $\langle Y \rangle$, are smoother than the reference field $Y$ (Figure~\ref{fig:logk_training}), but the realizations from the posterior ensemble exhibit more realistic features (right column in Figure~\ref{fig:logk_esmda}). Regardless of the forward model used (the only difference in these two experiments), our data assimilation framework yields consistent predictions of $\sigma_Y$ (middle column in  Figure~\ref{fig:logk_esmda}). It is small throughout most of the domain, indicating the reduced uncertainty in the estimation of hydraulic conductivity $K(\mathbf x)$ due to assimilation of the concentration and head measurements. The maximum values of $\sigma_Y$ and, hence, the largest predictive uncertainty in the $K(\mathbf x)$ estimation, are along the interface between the high- and low-conductivity regions. This finding suggests that the model predictions of hydraulic head and solute concentration are \textcolor{blue}{not significantly affected by} the changes in hydraulic conductivity in that domain; it reaffirms the conclusion of the sensitivity analysis of the relative importance of uncertainties in the spatial arrangement of hydrofacies and their hydraulic conductivities~\cite{winter-2006-multivariate}. \textcolor{blue}{Figure~\ref{fig:hist_std_logk} shows the histograms of the standard deviation $\sigma_Y$ on all pixels for CAAE-ESMDA AND CAAE-DenseED-ESMDA. The standard deviation of the ensemble for the latter one is overall slightly higher than that of the CAAE-ESMDA.}

\begin{figure}[htbp]
    \centering
    \includegraphics[width=0.8\textwidth]{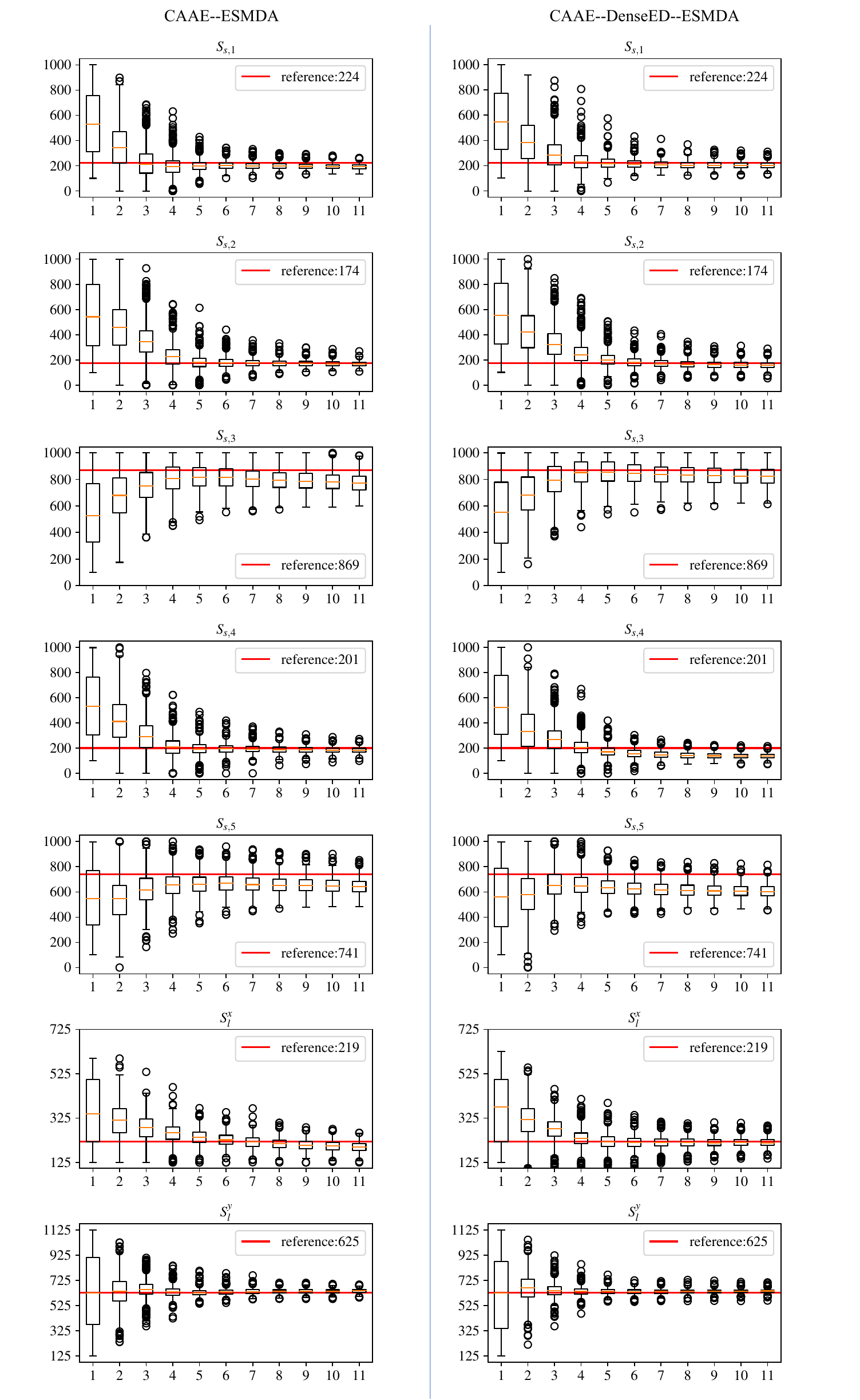}
    \caption{Boxplots of the ensembles for the contaminant release terms, $\mathbf S = (\mathbf S_\text{l},\mathbf S_\text{s})$ with $\mathbf S_\text{l} = (S_\text{l}^x, S_\text{l}^y)^\top$ and $\mathbf S_\text{s} = (S_{\text s,1},\dots,S_{\text s,5})^\top$, and their confidence intervals.\textcolor{blue}{Figures on the left column are obtained via the CAAE-ESMDA inversion with the PDE-based. forward model, the ones on the right column are obtained via CAAE-DenseED-ESMDA, which use a DenseED CNN surrogate.} These quantities are plotted as function of the ESMDA iterations and contrasted with their reference values (horizontal lines). The source location $\mathbf S_\text{l}$ is in m; and the contaminant release strength in each of the five stress periods, $\mathbf S_\text{s}$, is in g/m$^3$.}
    \label{fig:boxplot_source}
\end{figure}

The same inversion experiments yield estimates of the contaminant release history $\mathbf S$, which are shown in Figure~\ref{fig:boxplot_source}\textcolor{blue}{\st{ and Figure ...}} for CAAE-ESMDA and CAAE-DenseED-ESMDA, respectively. Regardless of the forward model used, our inversion algorithm accurately estimates the release strength during stress periods 1 and 2 ($S_{\text s, 1}$, $S_{\text s, 2}$); for period 4 ($S_{\text s,4}$), the discrepancy between the two experiments indicates the impact of the DenseED surrogate model error, which can be reduced by improving the DenseED training process; the estimates are close to their reference values and have tight 95\% confidence intervals. At the same time, the estimates of the source strength during stress periods 3 and 5 ($S_{\text s, 3}$ and $S_{\text s, 5}$) \textcolor{blue}{\st{fail to converge to their reference values and exhibit large error bars} although are not as close to the reference values, still encompass them within the 95\% confidence intervals.  Another potential cause for this deviation is the magnitude of the reference value: we observe that the error for the assimilated source terms is larger when the true reference value is large, not only in this experiment, but also in the two extra experiments in the supplemental material.} The divergence of $S_{\text s, 3}$ and $S_{\text s, 5}$ occurred in both experiments, \textcolor{blue}{\st{with or}} without the surrogate forward DenseED model, implying that the surrogate model error is not the source of the uncertainty in these two parameters. Additionally, since the $R^2$ of the prediction with CAAE is as high as $0.96$, we rule out CAAE as the source of this uncertainty and claim that this uncertainty is mostly due to the ill-posedness of the problem caused by the sparse and noisy measurements. The two assimilation strategies yield very similar estimates of the contaminant release location, $\mathbf S_\text{l} = (S_\text{l}^x, S_\text{l}^y)^\top$; the estimates of both quantities have tight confidence intervals, but the estimated value of $S_\text{l}^x$ for CAAE-DenseED-ESMDA inversion lies slightly farther from the reference value than that of CAAE-ESMDA. 

% \begin{figure}[htbp]
%     \centering
%     \includegraphics[width=0.7\textwidth]{fig9_boxplot-960-surro}
%     \caption{ Boxplots of the ensembles for the contaminant release terms, $\mathbf S = (\mathbf S_\text{l},\mathbf S_\text{s})$ with $\mathbf S_\text{l} = (S_\text{l}^x, S_\text{l}^y)^\top$ and $\mathbf S_\text{s} = (S_{\text s,1},\dots,S_{\text s,5})^\top$, and their confidence intervals, obtained via the CAAE-ESMDA inversion with the DenseED CNN surrogate. These quantities are plotted as function of the ESMDA iterations and contrasted with their reference values (horizontal lines). The source location $\mathbf S_\text{l}$ is in m; and the contaminant release strength in each of the five stress periods, $\mathbf S_\text{s}$, is in g/m$^3$.}
%     \label{fig:boxplot_source_surro}
% \end{figure} 

% \subsection{Comparison of Computational Costs}

The computational costs of CAAE-ESMDA with the PDE-based forward model and its counterpart with the DenseED CNN surrogate are shown in Table~\ref{tab:comp_cost}. CAAE-ESMDA with the PDE-based model ran on CPU, while the DenseED CNN surrogate was trained and simulated on GPUs provided by \texttt{GoogleColab}.  \textcolor{blue}{For clearer presentation of the error of the neural networks, we summarized the $R^2$ in Table~\ref{tab:R2_CAAE_DenseED}.}
In both cases, ESMDA consists of $N_\text{e} = 960$ samples in each ensemble and $10$ iterations are performed, resulting in $N_\text{sum} = N_\text{e}\times(10+1) = 10560$ forward model runs. Overall, CAAE-DenseED-ESMDA is one order of magnitude faster than CAAE-ESMDA with the PDE-based forward model.

%%%%%%%%%%%%%%%% Table of running time.
\begin{table}[htbp]
\centering
\caption{ Total run time of the CAAE-ESMDA, $T_\text{run}$, includes the costs of the PDE-based forward model and its CNN surrogate. The average run-time per sample, $T_\text{ave}$, is defined as $T_\text{ave} = (T_\text{run} + T_\text{dataset} + T_\text{train}^\text{DenseED}) / N_\text{sum}$, where $T_\text{dataset}$ is the time for obtaining the training and testing data sets, and $T_\text{train}$ is the CNN training time. CAAE parameterization is used in both cases, the training time is $18678.23$, the running time of CAAE is negligible in both data assimilation strategies. All times are in seconds.} 
\begin{tabular}{lrrrrr | r}
\hline
   & $N_\text{sum}$ & $T_\text{run}$ & $T_\text{dataset}$ &  $T_\text{train}^\text{DenseED}$ & $T_\text{ave}$ & \textcolor{blue}{$T_\text{train}^\text{CAAE}$}\\
\hline
  CAAE-ESMDA  & $10560$ & $388200.0$ & $0.0$ & $0.0$ & $36.8$ & \textcolor{blue}{$18678.23$}  \\
  CAAE-DenseED-ESMDA  & $10560$ &  $1893.9$ & $34922.0$ & $9439.2$ & $4.4$ &  \textcolor{blue}{$18678.23$}    \\
\hline
\end{tabular}
\label{tab:comp_cost}
\end{table}

%%%%%%%%%%%%%%%% Table of R2 coefficient of determination.
\begin{table}[htbp]
\centering
\caption{\textcolor{blue}{$R^2$ coefficient summary of the CAAE and DenseED. CAAE is trained with 23000 conductivity realizations. The evaluation of $R^2(\mathbf{k, \hat{k}})$ is through comparing the concentration and hydraulic head field of 150 realizations with the reconstructed conductivity fields and the original conductivity fields.  DenseED is trained with 800 realizations of different conductivity field and release histories. The evaluation of $R^2$ is through comparing the concentration and hydraulic head field of the same 150 realizations.}} 
\begin{tabular}{lrrr}
\hline
   & $N_\text{train}$ & $N^\text{simu}_\text{eval}$ & $R^2$ \\
\hline
  CAAE  & 23000 & 150 & 0.96\\
  DenseED  & 800 & 150 &  0.79\\
\hline
\end{tabular}
\label{tab:R2_CAAE_DenseED}
\end{table}

\section{Conclusions\textcolor{blue}{\st{and Discussion}}}
\label{sec:concl}

We proposed an CAAE-DenseED-ESMDA algorithm to infer  the statistics of both aquifer properties (e.g., hydraulic conductivity) and contaminant release history from sparse and noisy observations of hydraulic head and solute concentration. The algorithm relies on \textcolor{blue}{CAAE} to obtain a low-dimensional representation of the high-dimensional discretized conductivity field (and, if necessary, other spatially distributed input parameters); deploys a DenseED CNN surrogate of the PDE-based transport model to accelerate the forward runs; and adopts ESMDA to solve the inverse problem. The algorithm's computational efficiency is such that it enables one to handle three-dimensional problems. 

\textcolor{blue}{We also provide another two sets of CAAE-DenseED-ESMDA experiments to demonstrate the inversion with different conductivity fields and release history terms. These two experiments are gathered in Section~\ref{sec:append_B}. The ensembles for contaminant release history terms in both experiment converge to the true values with at most $11.75\%$ deviation. We observe high uncertainty and slightly high deviation from the true value on the terms that has higher absolute value. In terms of the conductivity field, the magnitudes of the standard deviation in these experiments are also comparable to the experiment presented in the main text. The structure or characteristics of the conductivity field results is also consistent with the ones shown in the main text.}

\textcolor{blue}{Methodologically, deployment of CAAE-DenseED-ESMDA allows one to investigate questions, such as measuring the data assimilation accuracy versus the ensemble size or designing a network of observation wells, that cannot be answered with CAAE-ESMDA with the PDE-based forward model, whose computational cost might be prohibitive.} To demonstrate the salient features of our inversion methodology, we conduct a series of numerical experiments. They deal with flow and transport in a three-dimensional heterogeneous aquifer with uncertain hydraulic conductivity field; our goal is to estimate the latter, and the contaminant release history, from the measurements of hydraulic head and contaminant concentration collected in a few observation wells. These numerical experiments lead to the following conclusions.

\begin{enumerate}
    \item The CAAE-DenseED-ESMDA inversion framework is capable of both identifying the contaminant release source and reconstructing a three-dimensional hydraulic conductivity field from sparse (in space and time) and noisy measurements of solute concentration and hydraulic head.
    \item The CAAE-ESMDA inversion, with or without the DenseED CNN surrogate of the PDE-based forward model, yields estimates of the contaminant release strength that differ from the reference values by up to $30.42\%$. 
    That can be attributed to the imperfect reconstruction of hydraulic conductivity field or relative insensitivity of the observed solute concentrations to the contaminant release strengths in each stress period (the inverse problem's ill-posedness). \textcolor{blue}{That being said, the ensembles for the release history terms all covered the groundtruth values, as an evidence of the capability of the proposed framework.}
    \item Deployment of the DenseED CNN surrogate within our CAAE-ESMDA inversion framework provides an order of magnitude speed up, while giving \textcolor{blue}{visually similar} estimates of the hydraulic conductivity field; it also increases the predictive uncertainty (posterior standard deviation) relative to that obtained via the CAAE-ESMDA inversion with the PDE-based model \textcolor{blue}{as shown in Figure~\ref{fig:hist_std_logk}. The increase of the uncertainty of the conductivity field results is not substantial considering that the largest value of $\sigma_{Y}$ does not differ much for two experiments, and the overlapping area of the histogram is big. Quantifying the relation between the $R^2$ of DenseED and the increased $\sigma_{Y}$ will required future experiments spanning several more sets of DenseED experiments.}  
    \item The computational efficiency of CAAE-DenseED-ESMDA, relative to that of CAAE--ESMDA with the high-fidelity PDE model, is mostly due to the use of GPUs for CNN-related computations, while the PDE solver for the flow and transport model (e.g., MODLFLOW and MT3DMS) utilizes CPUs.
    \item \textcolor{blue}{\st{Deployment of CAAE-DenseED-ESMDA allows one to investigate questions, such as measuring the data assimilation accuracy versus the ensemble size or designing a network of observation wells, that cannot be answered with CAAE-ESMDA with the PDE-based forward model, whose computational cost might be prohibitive.}}
\end{enumerate}

\section{\textcolor{blue}{Discussion}}
\label{sec:disc}

Although the flow and transport simulators, MODFLOW and MT3DMS, can be parallelized to run on multiple CPU cores, that is a much more arduous task than carrying out NN-related computations on GPUs available in Google-Colab or other cloud computing environments. The latter takes very little implementation effort and can be done on a personal computer. The advantage of our method largely depends on the feasibility of accessing GPU computing resources versus deploying multi-core parallelization with the physics-based forward model.

Our method can be extended to handle other unknown parameters, such as porosity or reaction rate constants, with no significant adjustment. For example, a spatially variable porosity field can be treated similarly to the permeability field in our experiment, i.e., CAAE parameterization can be used to re-parameterize the porosity field, and the corresponding latent variable can be inferred by ESMDA. \textcolor{blue}{A parameter that is not} spatially-dependent can be added as an extra channel to the input of the CNN surrogate forward model. The corresponding ESMDA inversion part would be similar to that for other parameters. \textcolor{blue}{Though the implementation of our framework on another application will require nontrivial effort of tuning and might result in different level of uncertainty of the inversion results, the established framework itself is capable to accommodate the required changes.}

Our numerical experiments utilize CAAE for parameterization, which has two main benefits. First, it relieves the computational burden of the ESMDA inversion. In our experiments, CAAE reduced the number of the total parameters to be inferred from $19933$ to $931$. Since the computational cost of ESMDA is linear in the state size~\cite{evensen2019efficient}, the ESMDA inversion is accelerated by $\approx 20$ times. Second, CAAE is capable of capturing the channelized characteristics of a conductivity field as this prior information is integrated into the CAAE at the training stage. The ESMDA inversion without CAAE might yield a conductivity field that either has a lower resolution or loses the sharp edges of the channels \cite{kang2021hydrogeophysical}. Hence, even if the computation savings of CAAE (the first benefit) do not outweigh its training cost, the presence of channels calls for its use (the second benefit).

The integration of the three distinct components into a single inversion framework has its limitations. \textbf{CAAE:} The loss of fine features of a conductivity field is unavoidable in reduced-order modeling (see, e.g.,  Figure~\ref{fig:release_candi}). It is not trivial to perform ESMDA without CAAE because ESMDA assumes Gaussianity of an unknown parameter, which is invalid for the original conductivity field. This undermines the veracity of the inversion procedure with or without CAAE. The time for training CAAE neural networks in similar studies is reported as \textcolor{blue}{$1.7$ and $2.7$ hours for 2D fields with discretization $32 \times 64$ and $40 \times 80$ in \cite{mo2019integration,kang2021hydrogeophysical}; for 3D fields with discretization $6 \times 32 \times 64$, the training time reported in \cite{mo2019integration} is $13.1$ hours, for our experiment, it took $5.2$ hours. The CAAE training time} might negate the computational gain from the parameterization in the inversion process. \textbf{ESMDA:} Bench-marking against other more accurate inversion methods such as MCMC requires exhaustive simulations given the large number of unknown parameters. \textcolor{blue}{\st{This precludes us from separating the error caused by ESMDA inversion from the overall inversion error.}}. Since ESMDA requires an unknown parameter to be Gaussian, its application to a typical subsurface problem is impossible without a parameterization. \textcolor{blue}{One way to estimate the effect of the chosen setting of ESMDA (number of samples in each ensemble, number of iterations, etc.) is perform an analysis of the inversion accuracy versus different settings of ESMDA. We leave it as a future exploration.} \textbf{DenseED:} The inversion accuracy of CAAE-DenseED-ESMDA is related to the quality of both a CAAE parameterization and a DenseED surrogate model. There is a trade-off between the computational time and accuracy unless neural network surrogates can be improved such that they would not require many hours to train or require significantly fewer simulations than the ESMDA procedure \cite{song-2021-transfer}. \textcolor{blue}{Similar to our note about analyzing the setting of ESMDA, for DenseED surrogate model, a set of experiment involving different number of training samples, details of the surrogate model architecture, training epochs, etc. could reveal the upper bound of the quality of DenseED, and provide more insight of the robustness of the framework. We leave it as a future research experiment as well.} Another limitation of our method stems from the fixed input-output structure of the DenseED surrogate forward model. Once the DenseED is trained, it does not generalize to produce predictions at arbitrary time, limiting the contaminant release history to predefined times.

\section*{Acknowledgements}
ZZ and DT were supported in part by National Science Foundation grant EAR-2100927, and by a gift from Total. NZ acknowledges  support from ARPA-E, award \# DE-AR0001204. There are no data sharing issues since all of the numerical information is provided in the figures produced by solving the equations in the paper.  We reused the channelized conductivity field data from the open sourced dataset in \url{https://github.com/GAIA-UNIL/trainingimages}, and simulated the contaminant transport processes with MODFLOW and MT3DMS. These data and the source code are available at https://doi.org/10.5281/zenodo.6443086.

% \begin{comment}
\section*{Appendix A}
The following sections discuss the details of the DNNs used in this study: CAAE for the parameterization of the hydraulic conductivity field, and the DenseED surrogate model predicting the groundwater flow and the contaminant transport.
\subsection*{CAAE}
\begin{figure}[htbp]
    \centering
    \includegraphics[width=\textwidth]{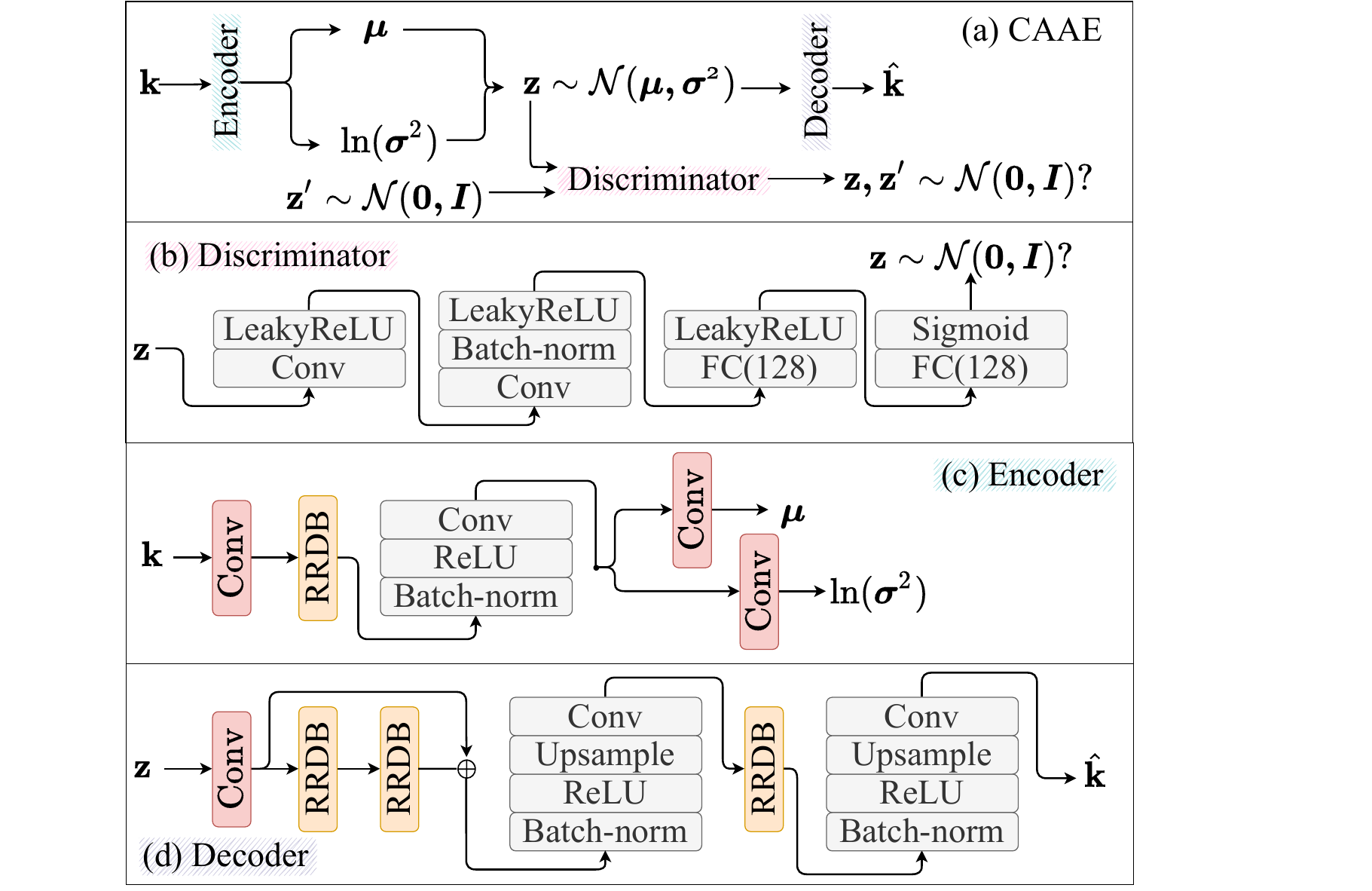}
    \caption{\em (a) CAAE, (b) Discriminator, (c) Encoder, and (d) Decoder. The CAAE (a) consists of an encoder (c) and a decoder (d), the discriminator (b) is trained as well to enforce the distribution of the low-dimensional latent variable $\mathbf{z}$.  ``RRDB'' blocks are depicted in Figure~\ref{fig:RRDB}, with the slope parameter in ``LeakyReLU'' being $0.2$ in this study. $\oplus$ denotes element-wise summation.}
    \label{fig:CAAE_arch}
\end{figure}

\begin{figure}[htbp]
    \centering
    \includegraphics[width=0.6\textwidth]{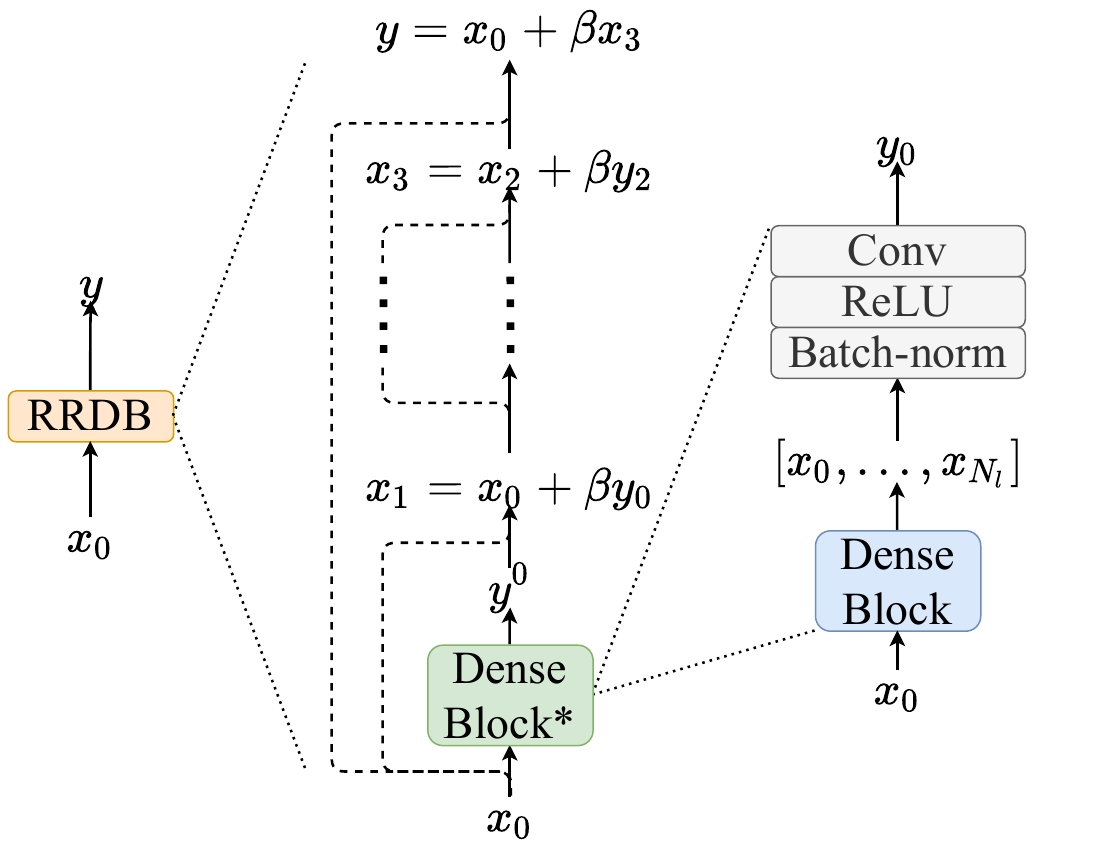}
    \caption{\em RRDB block structure. Each RRDB block used in this study consists of $N_{DB}=3$ dense blocks, with the internal layer number of each dense block being $N_l=5$. The dense block is illustrated in (a) in Figure~\ref{fig:dense_block}.}
    \label{fig:RRDB}
\end{figure}

As briefly introduced in Section~\ref{sec:CAAE}, the CAAE consists of three networks: an encoder ($\mathcal G$), a decoder ($\text{De}$), and a discriminator ($\mathcal D$). The workflow of these three networks and the architecture of each network are shown in Figure~\ref{fig:CAAE_arch}. The residual-in-residual dense block (RRDB)  used in the encoder and the decoder is illustrated in Figure~\ref{fig:RRDB}, in which the dense block is reused in the DenseED surrogate model as well, and is shown in Figure~\ref{fig:dense_block}. The dimensions of the internal layers of the encoder and the decoder are shown in Table~\ref{tab:CAAE_dimensions}. Batch normalization (BN)~\cite{ioffe2015batch}, three-dimensional convolutional operations (Conv)~\cite{goodfellow2016deep}, Sigmoid, ReLU and LeakyReLU nonlinear activation functions~\cite{he2015delving} are used in these neural networks; ``FC(128)'' denotes a fully-connected layer with the output vector length being $128$, ``Upsample'' layer doubles the size of the input feature maps with the nearest upsampling method.

%%%%%%%%%%%%%%%% Table of DenseED dimensions.
\begin{table}[htbp]
\centering
\caption{\em Dimension of the internal layer outputs of the encoder and decoder in CAAE. The encoder outputs $\boldsymbol{\mu}$ and $\ln \boldsymbol{\sigma}^2$. $N_{DB}$ denotes the number of dense blocks in a residual-in-residual block, shown in Figure~\ref{fig:RRDB}.} 
\begin{tabular}{lll}
\hline
\multicolumn{3}{c}{Encoder}\\ 
\hline
   Layers & Number of features $C_f$ & Resolution $W\times H \times D$\\
\hline
   Input: $\mathbf{k}$ & $1$ & $81\times 41 \times 6$ \\
   Conv & $48$ & $41\times 21 \times 3$ \\
   RRDB, $N_{DB}=3$ & $48$ & $41\times 21 \times 3$\\
   BN-ReLU-Conv & $48$ & $41\times 21 \times 3$\\
   Conv: $\boldsymbol{\mu}$ & $2$ & $21\times 11 \times 2$ \\
   Conv: $\ln{\boldsymbol{\sigma}^2}$ & $2$ & $21\times 11 \times 2$ \\
\hline
\multicolumn{3}{c}{Decoder}\\
\hline
   Layers & Number of features $C_f$ & Resolution $W\times H \times D$\\
\hline
   Input: $\mathbf{z}$ & $2$ & $21\times 11 \times 2$ \\
   Conv & $48$ & $21\times 11 \times 2$ \\
   RRDB, $N_{DB}=3$ & $48$ & $21\times 11 \times 2$ \\
   RRDB, $N_{DB}=3$ & $48$ & $21\times 11 \times 2$ \\
   BN-ReLU-UP-Conv & $48$ & $41\times 21 \times 4$\\
   RRDB, $N_{DB}=3$ & $48$ & $41\times 21 \times 4$ \\
   BN-ReLU-UP-Conv: $\hat{\mathbf{k}}$ & $1$ & $81\times 41 \times 6$\\
\end{tabular}
\label{tab:CAAE_dimensions}
\end{table}

\begin{figure}[htbp]
    \centering
    \includegraphics[width=0.9\textwidth]{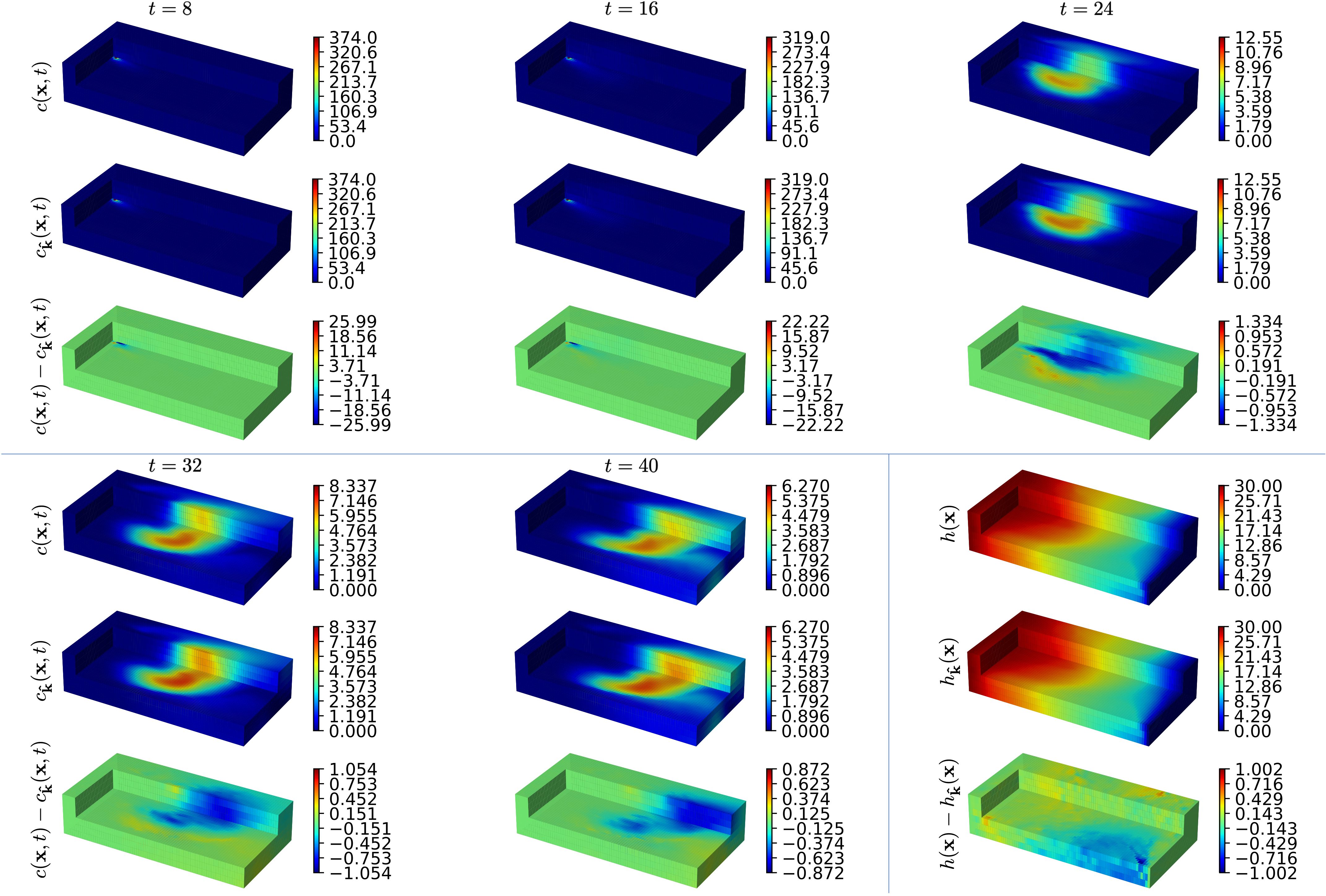}
    \caption{\em \textcolor{blue}{Prediction of the solute concentration obtained with the original training set of conductivity field, $c(\mathbf{x}, t)$, and with the CAAE reconstructed conductivity field,  $c_\mathbf{\hat{k}}(\mathbf{x}, t)$. Also shown are the corresponding predictions of the hydraulic head, $h(\mathbf{x})$ and $h\mathbf{\hat{k}}(\mathbf{x})$; and the difference between these two types.}}
    \label{fig:simu_w_fake_KD}
\end{figure}

\textcolor{blue}{Figure~\ref{fig:simu_w_fake_KD} shows an example of the PDE-based simulation with the reconstructed conductivity field. This figure can visually show the effect of using CAAE on the predicted fields, as a supplemental material to the $R^2$ of the CAAE. This figure and Figure~\ref{fig:surro_nn_3dkd} use the same set of release history and conductivity field input.}

\subsection*{DenseED}
\begin{figure}[htbp]
    \centering
    \includegraphics[width=\textwidth]{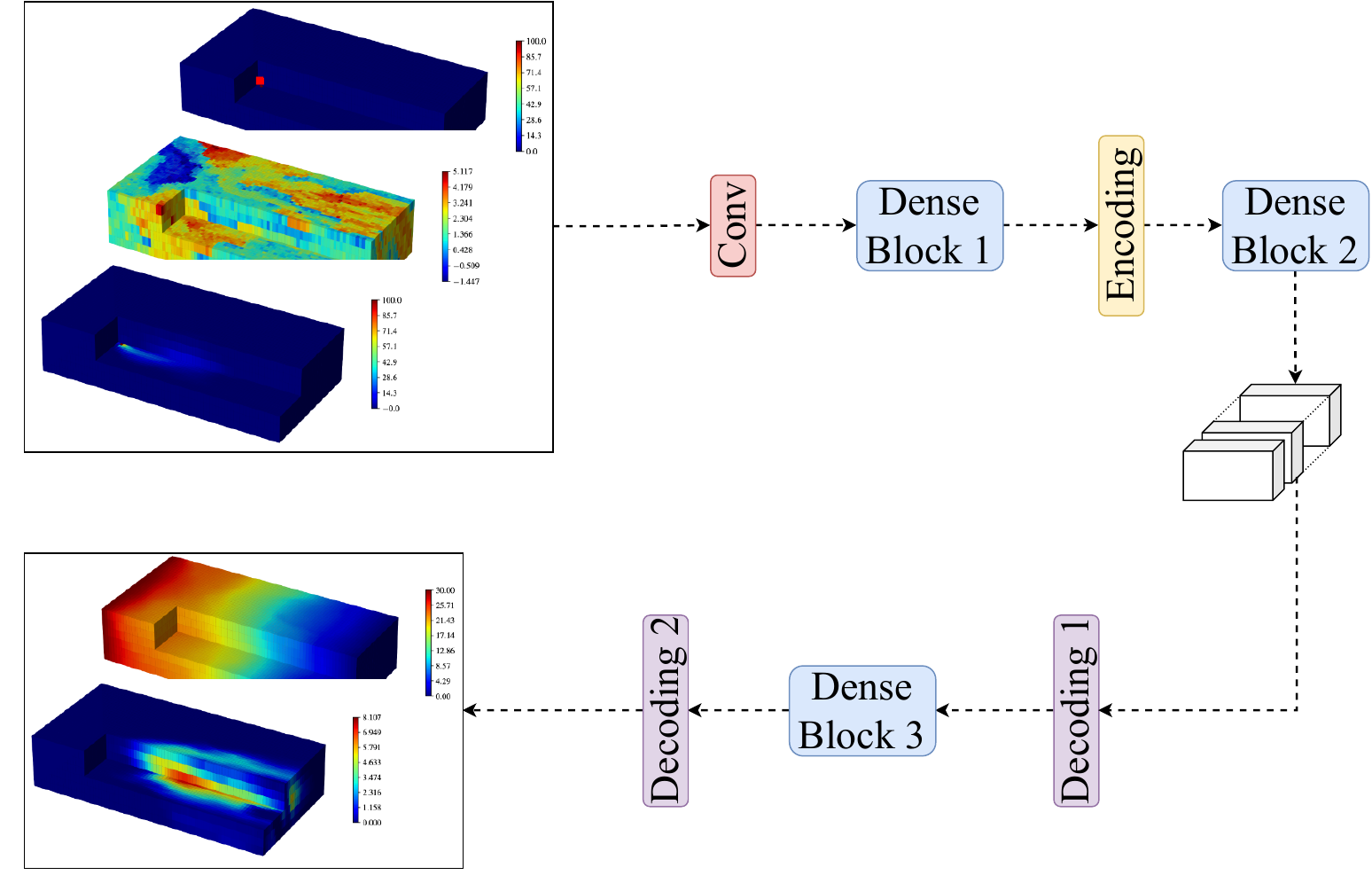}
    \caption{\em Dense encoder decoder (DenseED) architecture with three dense blocks. The cubes as the output of  ``Dense Block 2'' represent the encoded coarse high-level features.}
    \label{fig:denseED}
\end{figure}

\begin{figure}[htbp]
    \centering
    \includegraphics[width=0.9\textwidth]{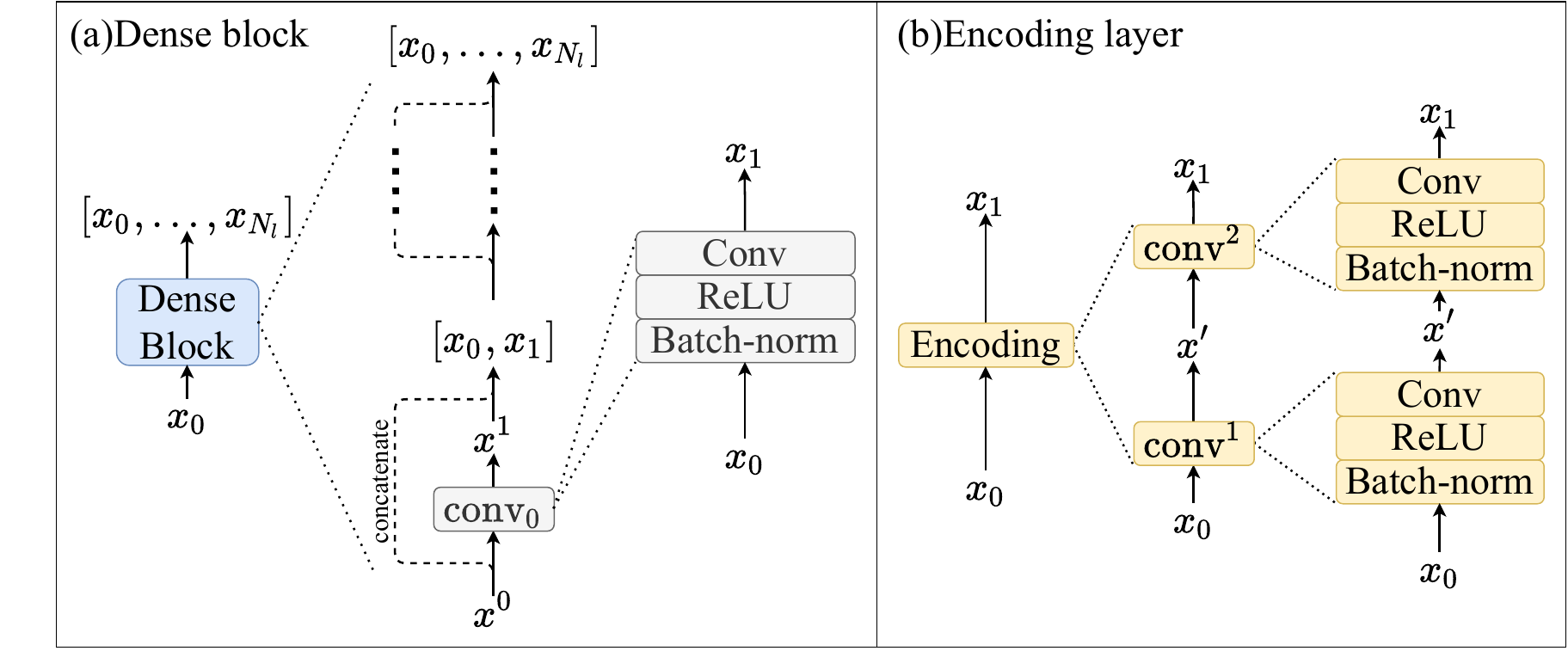}
    \caption{\em (a)A dense block with $N_l$ internal layers. (b)Encoding layer structure in Figure~\ref{fig:denseED}. The feature number of $x^0$ is halved first, then the size of the features are halved: $x^0 \in \mathbb{R}^{C\times W\times H\times D}, x'\in \mathbb{R}^{\frac{1}{2}C\times W\times H\times D}, x^1 \in \mathbb{R}^{\frac{1}{2}C\times \frac{1}{2}W\times \frac{1}{2}H\times \frac{1}{2}D} $.}
    \label{fig:dense_block}
\end{figure}

The architecture of the DenseED in this study is shown in Figure~\ref{fig:denseED}. The DenseED neural network structure consists of three main sub-structures: dense blocks, encoding layers, and decoding layers. The structure of a dense block is illustrated in (a) in Figure~\ref{fig:dense_block}. An encoding layer is shown in (b) in Figure~\ref{fig:dense_block}, with which both the feature number and the size of the features are halved: $x^0 \in \mathbb{R}^{C\times W\times H \times D}, x'\in \mathbb{R}^{\frac{1}{2}C\times W\times H \times D}, x^1 \in \mathbb{R}^{\frac{1}{2}C\times \frac{1}{2}W\times \frac{1}{2}H\times \frac{1}{2}D} $. This figure can represent a decoding layer as well, with the feature number halved, the size doubled: $x^0 \in \mathbb{R}^{C\times W\times H\times D}, x'\in \mathbb{R}^{\frac{1}{2}C\times W\times H\times D}, x^1 \in \mathbb{R}^{\frac{1}{2}C\times 2W\times 2H\times 2D} $.  In addition to these three main elements, the size of the features are first halved with the very first $\text{Conv}$ layer. The last decoding layer maps the feature number to that of the output. The dimensions of the outputs from each block are shown in Table~\ref{tab:DenseED_dimensions}.

%%%%%%%%%%%%%%%% Table of DenseED dimensions.
\begin{table}[htbp]
\centering
\caption{\em Dimension of the internal layer outputs of the DenseED network. $N_l$ is the number of the internal layers in a dense block shown in Figure~\ref{fig:dense_block}.} 
\begin{tabular}{lll}
\hline
   Layer & Number of features $C_f$ & Resolution $W\times H\times D$\\
\hline
   Input: $(c(\mathbf{x},t), \mathbf{k}, S(\mathbf{x},t))$ & $3$ & $81\times 41 \times 6$ \\
   Conv & $48$ & $41\times 21 \times 3$ \\
   Dense Block 1, $N_l=3$ & $192$ & $41\times 21 \times 3$\\
   Encoding & $96$ & $21\times 11 \times 2$ \\
   Dense Block 2, $N_l=6$ & $384$ & $21\times 11 \times 2$\\
   Decoding 1& $192$ & $41 \times 21 \times 3$ \\
   Dense Block 3, $N_l=3$ & $336$ & $41\times 21 \times 3$\\
   Decoding 2: $(c(\mathbf{x},t+\Delta t), h(\mathbf{x}))$& $2$ & $81 \times 41 \times 6$ \\
\end{tabular}
\label{tab:DenseED_dimensions}
\end{table}

\newpage

\section*{Appendix B}
\label{sec:append_B}
\subsection*{Two more sets of CAAE-DenseED-ESMDA experiments}
\begin{figure}[htbp]
    \centering
    \includegraphics[width=0.7\textwidth]{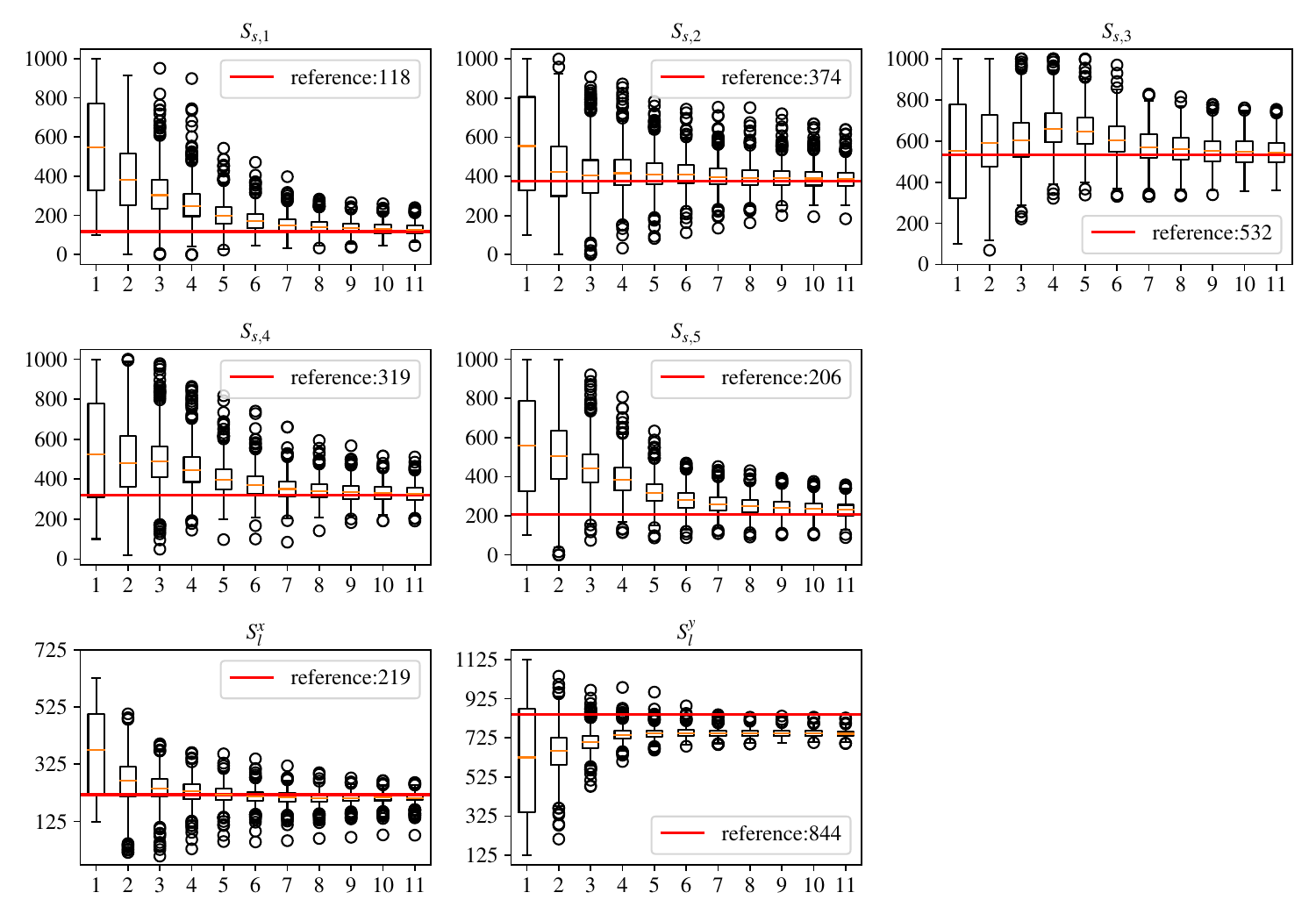}
    \caption{\em Boxplots of the ensembles for the contaminant release terms, $\mathbf S = (\mathbf S_\text{l},\mathbf S_\text{s})$ with $\mathbf S_\text{l} = (S_\text{l}^x, S_\text{l}^y)^\top$ and $\mathbf S_\text{s} = (S_{\text s,1},\dots,S_{\text s,5})^\top$, and their confidence intervals, obtained via the CAAE-DenseED-ESMDA inversion with the PDE-based forward model. These quantities are plotted as function of the ESMDA iterations and contrasted with their reference values (horizontal lines). The source location $\mathbf S_\text{l}$ is in m; and the contaminant release strength in each of the five stress periods, $\mathbf S_\text{s}$, is in g/m$^3$.}
    \label{fig:boxplot_source_surro_set2}
\end{figure}

\begin{figure}[htbp]
    \centering
    \includegraphics[width=\textwidth]{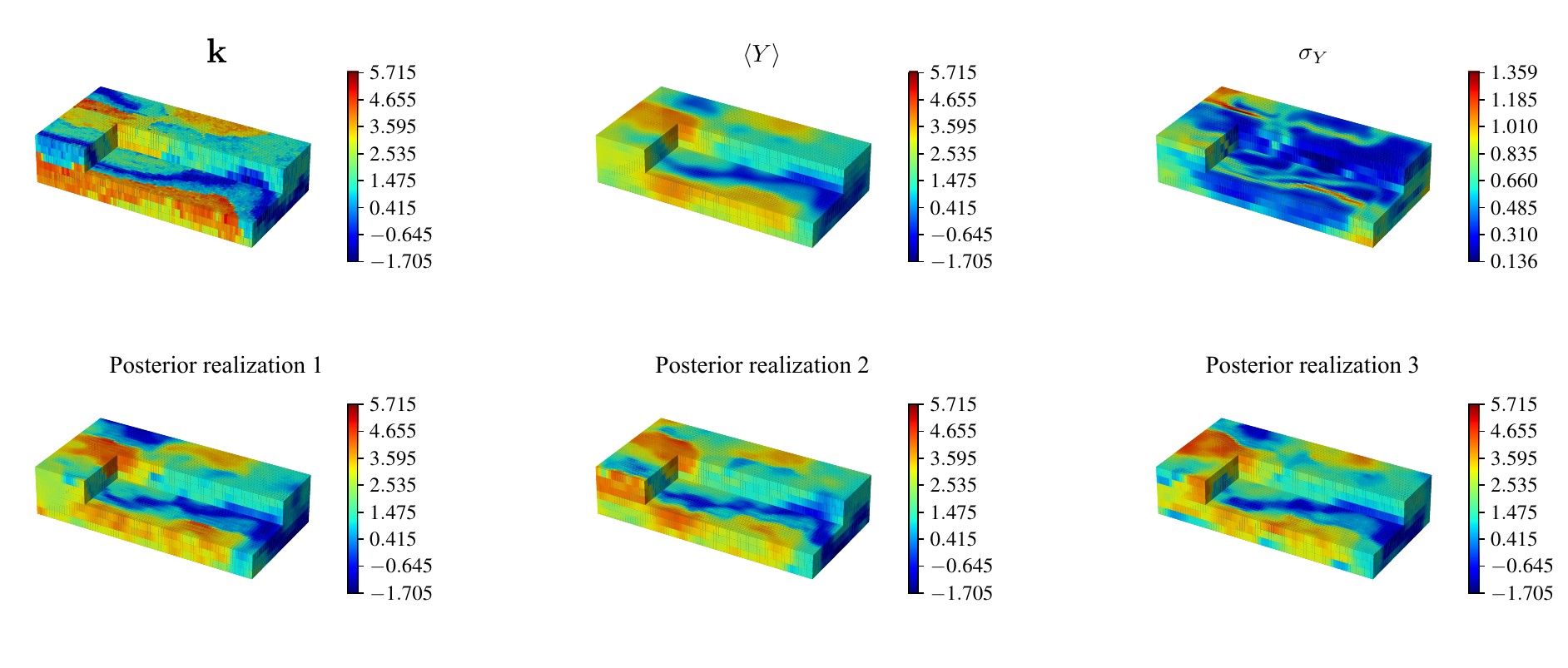}
    \caption{\em True conductivity field $\mathbf k$, posterior mean ($\langle Y \rangle$, left column) and standard deviation ($\sigma_{Y}$, middle column) of the log-conductivity field $Y(\mathbf x)$ obtained upon assimilation of concentration and head measurements from the dense observation network. These statistics are obtained via our inversion algorithm CAAE-DenseED-ESMDA. Also shown are representative three realizations from the resulting posterior ensemble (bottom row). 
    }
    \label{fig:logk_esmda_set2}
\end{figure}

\begin{figure}[htbp]
    \centering
    \includegraphics[width=0.7\textwidth]{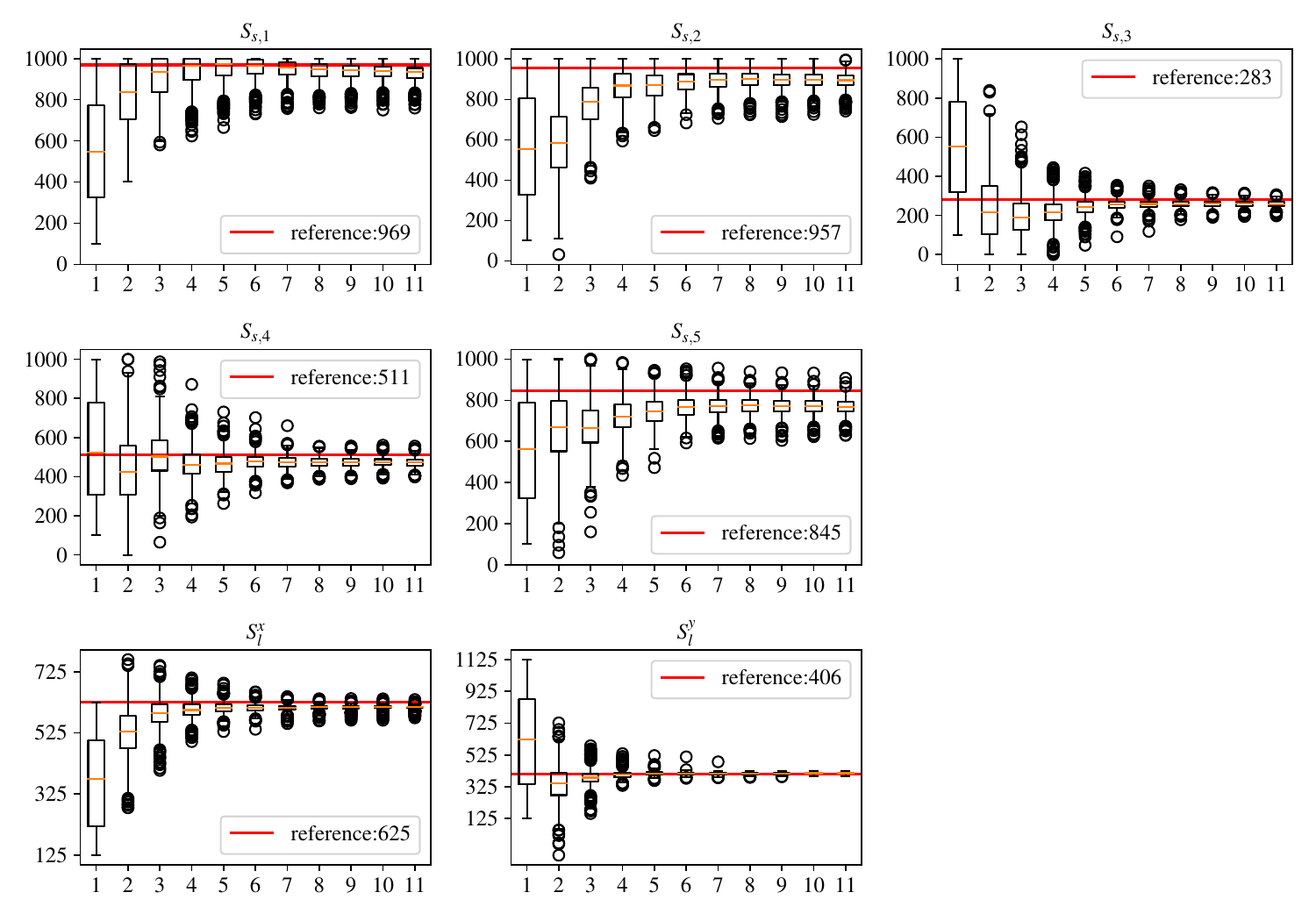}
    \caption{\em Boxplots of the ensembles for the contaminant release terms, $\mathbf S = (\mathbf S_\text{l},\mathbf S_\text{s})$ with $\mathbf S_\text{l} = (S_\text{l}^x, S_\text{l}^y)^\top$ and $\mathbf S_\text{s} = (S_{\text s,1},\dots,S_{\text s,5})^\top$, and their confidence intervals, obtained via the CAAE-DenseED-ESMDA inversion with the PDE-based forward model. These quantities are plotted as function of the ESMDA iterations and contrasted with their reference values (horizontal lines). The source location $\mathbf S_\text{l}$ is in m; and the contaminant release strength in each of the five stress periods, $\mathbf S_\text{s}$, is in g/m$^3$.}
    \label{fig:boxplot_source_surro_set3}
\end{figure}

\begin{figure}[htbp]
    \centering
    \includegraphics[width=\textwidth]{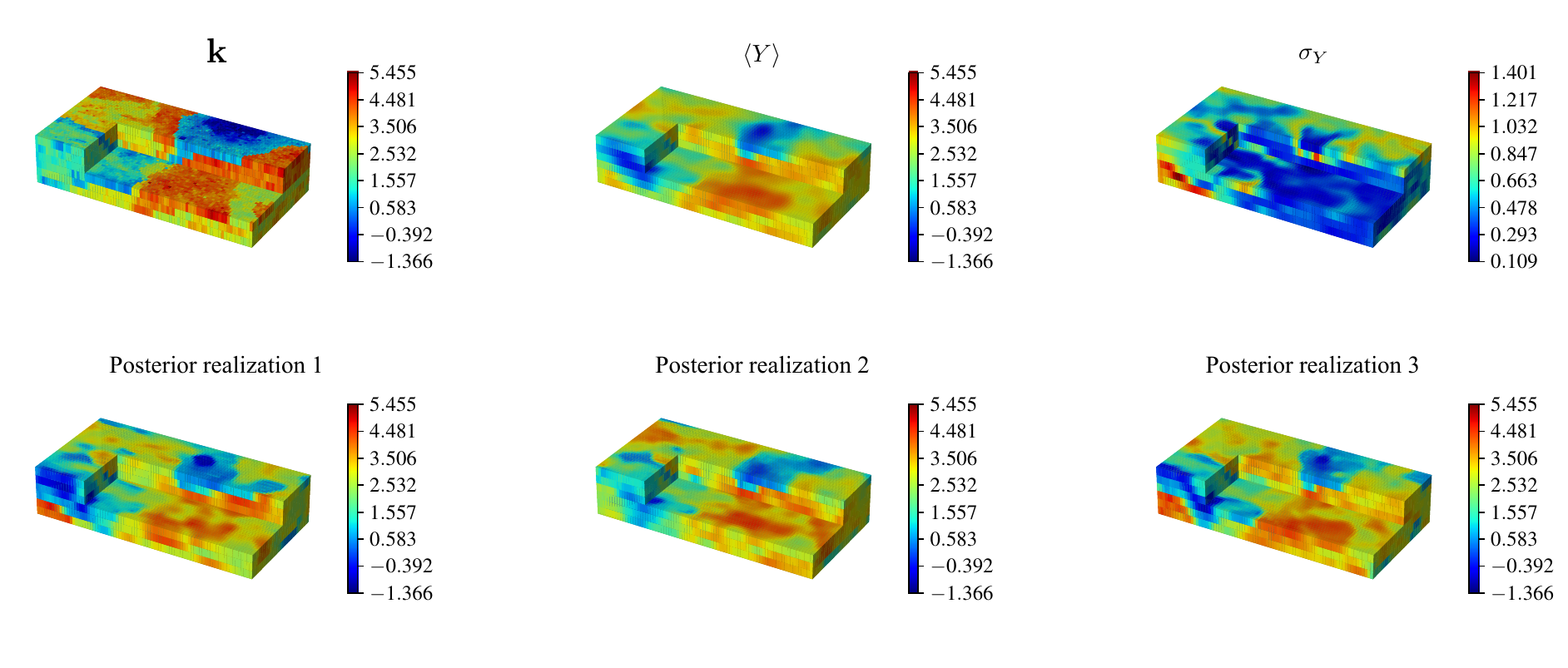}
    \caption{\em True conductivity field $\mathbf k$, posterior mean ($\langle Y \rangle$, left column) and standard deviation ($\sigma_{Y}$, middle column) of the log-conductivity field $Y(\mathbf x)$ obtained upon assimilation of concentration and head measurements from the dense observation network. These statistics are obtained via our inversion algorithm CAAE-DenseED-ESMDA. Also shown are representative three realizations from the resulting posterior ensemble (bottom row). 
    }
    \label{fig:logk_esmda_set3}
\end{figure}

In this section, we show another two sets of CAAE-DenseED-ESMDA inversion experiments with different release history and conductivity fields from what was shown in Section~\ref{sec:results}.
The results of the second set of experiments are shown in Figures~\ref{fig:boxplot_source_surro_set2} and~\ref{fig:logk_esmda_set2}.
The results of the third set of experiments are shown in Figures~\ref{fig:boxplot_source_surro_set3} and~\ref{fig:logk_esmda_set3}. These two sets of results are both obtained with our proposed CAAE-DenseED-ESMDA framework. The reference (true) value of the conductivity field and the release history are plotted in those figures as well. Both results show good quality of contamination strength identification, with the maximum discrepancy of the release strength being $11.75\%$ in the second experiment, and $9.05\%$ in the third experiment. The assimilation for the second experiment of the release location achieves similar performance with the experiment in Section~\ref{sec:results}, and that of the third experiment was the best among all experiments. The reconstructed conductivity field in Figures~\ref{fig:logk_esmda_set2} and~\ref{fig:logk_esmda_set3} both captured part of the true fields, yet with high uncertainty, which might be the effect of CAAE reconstruction error, forward surrogate model DenseED error, and the uncertainty arising from the sparse and noisy measurements.

\section*{Appendix C}
\label{sec:append_c}

\textcolor{blue}{We summarize the constants and discretization of the fields in Table~\ref{tab:const_discr}.}

\begin{table}[H]
    \centering
    \caption{Constants and dimensions of the experiment settings.}
    \begin{tabular}{l l l}
        Term & Representation & Value \\
        \hline
        $N_\text{re}$ & number of release period & $5$ \\
        $M$ & number of measurement location & $144$ \\
        $I$ & number of measurement time for concentration & $10$ \\
        dim($\mathbf{k}$) & dimension of conductivity field & $81 \times 41 \times 6$ \\
        $N_d$ & number of total measurements & $M(I+1) = 1584$ \\
        dim($\mathbf{z}$) & dimension of latent $\mathbf{z}$ & $2 \times 2 \times 11 \times 21 $\\
        $N_m$ & number of unknown parameters & $N_\text{re}+2+2 \times 2 \times 11\times 21 = 931$\\
        $N_e$ & number of samples in an ESMDA ensemble & $960$ \\
        $N_a$ & number of ESMDA iteration & $10$ \\
    \end{tabular}
    \label{tab:const_discr}
\end{table}

% \end{comment}

% \bibliography{references}

\end{document}